%% file: main.tex
\documentclass[12pt]{article}

\usepackage{amsmath}
\usepackage{amsthm}
\usepackage{amssymb}
\usepackage{graphicx}
\usepackage[margin=1in]{geometry}
\usepackage{subfigure}
\usepackage{url}
\usepackage{multirow}
\usepackage{algorithm}
\usepackage{algpseudocode}
\usepackage{rotating}
\usepackage[table]{xcolor}

\title{PACOL: Poisoning Attacks Against Continual Learners}
\author{Huayu Liu$^1$ and Gregory Ditzler$^2$\footnote{Gregory Ditzler was affiliated with the University of Arizona and Rowan University when this work was performed.}}
\date{
$^1$University of Arizona, Tucson, AZ 85721 USA \\
$^2$EpiSys Science, Philadelphia, PA 19128 USA
}

\begin{document}

\maketitle

\begin{abstract}
Continual learning algorithms are typically exposed to untrusted sources that contain training data inserted by adversaries and bad actors.  An adversary can insert a small number of poisoned samples, such as mislabeled samples from previously learned tasks, or intentional adversarial perturbed samples, into the training datasets, which can drastically reduce the model's performance. In this work, we demonstrate that continual learning systems can be manipulated by malicious misinformation and present a new category of data poisoning attacks specific for continual learners, which we refer to as {\em Poisoning Attacks Against Continual Learners} (PACOL). The effectiveness of labeling flipping attacks inspires PACOL; however, PACOL produces attack samples that do not change the sample's label and produce an attack that causes catastrophic forgetting. A comprehensive set of experiments shows the vulnerability of commonly used generative replay and regularization-based continual learning approaches against attack methods. We evaluate the ability of label-flipping and a new adversarial poison attack, namely PACOL proposed in this work, to force the continual learning system to forget the knowledge of a learned task(s). More specifically, we compared the performance degradation of continual learning systems trained on benchmark data streams with and without poisoning attacks. Moreover, we discuss the stealthiness of the attacks in which we test the success rate of data sanitization defense and other outlier detection-based defenses for filtering out adversarial samples.

\end{abstract}

\input{content/introduction}
\input{content/related_works}

\input{content/attacks}

\input{content/experiments}

\input{content/Conclusion}
\section*{Acknowledgement}
This work was supported by the Department of Energy \#DE-NA0003946, and the National Science Foundation CAREER \#1943552. Any opinions, findings, conclusions, or recommendations expressed in this material are those of the authors and do not necessarily reflect the sponsors' views. G. Ditzler was affiliated with the University of Arizona and Rowan University when this work was performed. 

\bibliographystyle{ieeetr}
\bibliography{bib/greg,bib/huayu}

\end{document}

%% file: content/introduction.tex
\section{Introduction}
Many information sources, including sensor networks, financial markets, social networks, and healthcare monitoring, are data streams arriving sequentially over time. Consequently, the learning procedures of intelligent systems take place in real-time, with partial data and without the capacity to store the entire data set~\cite{Ditzler2015CIM, Grossberg1988NN, Polikar2001TSMC}. With the rapid development of deep learning techniques, modern intelligent systems have adopted deep neural networks as their core computational model. Unfortunately, neural networks suffer catastrophic forgetting~\cite{mccloskey1989catastrophic} when they are tasked with learning from sequential or streaming data, which means a well-trained neural network can partially, or even entirely, forget previously learned knowledge. Data streams are typically sampled from changing environments and have limited or no historical data access. Note that this streaming paradigm is different from traditional offline settings of single-task learning where the entire dataset is sampled from an i.i.d. distribution. 
Continual learning, also called incremental learning and lifelong learning, is designed to retain the knowledge of the learned tasks by preserving the important model parameters for previous tasks. The goal is to prevent these parameters from experiencing large changes when new data become available. 
There are two core properties of a successful continual learning system: \emph{stability} and \emph{plasticity}~\cite{Grossberg1988NN}. \emph{Stability} is the property of a model to retain old knowledge, and \emph{plasticity} is the property that allows the model to learn from new data. A good continual learning system must maintain a healthy balance between \emph{stability} and \emph{plasticity} to achieve lifelong learning from massive data under the limitations of finite storage. 

On the other hand, learning on data streams increases the risk of the continual learning system's exposure to malicious data \cite{Li2021ACM}. 
The adversary can inject malicious data at training or testing time \cite{Biggio2017arxiv}. These types of malicious data injection at training and testing time are called poisoning and evasion attacks, respectively. 
Recent work has shown that neural networks are particularly vulnerable to data poisoning attacks~\cite{Biggio2012ICML, Xiao2012ECAI, Xiao2015ICML, Liu2021IS}. Adversaries can inject poisoning samples with manipulated labels or feature values (e.g., pixels for image data) into the training data that cause deleterious predictions and decrease the robustness of the model. Early studies focus mainly on poisoning offline datasets~\cite{Biggio2012ICML, Xiao2015ICML, munoz2017towards, mei2015using, Koh2022ML} where the model is trained on a fixed task. Recent progress shows that poisoning real-time data streams is considered a more practical scene where adversaries can interact with the training process and dynamically poison the data batches according to the model states. Carlini et al. showed that web-scale datasets could be poisoned for less than \$60 USD \cite{Carlini2023arxiv}. 
Poisoning attacks against online learning~\cite{pang2021accumulative, wang2018data, zhang2020online} have been proposed to dynamically poison each data batch when pre-trained neural networks are fine-tuned on sequentially captured real-time data. Compared to poisoning attacks against offline learning settings, online poisoning attacks craft poisoning samples according to the current model state, which is more flexible and can cause greater degradation to model accuracy. Moreover, poisoning attacks have been launched towards collaborative paradigms~\cite{tolpegin2020data,bagdasaryan2020backdoor}, such as federated learning~\cite{konevcny2016federated}. The distributed nature of federated learning gives rise to threats caused by malicious participants that sharing the model with distributed clients facilitates white-box accessibility to model parameters.

Unfortunately, while several data poisoning methods for online learning and federated have been widely studied, data poisoning in continual learning scenarios has received significantly less attention than their offline counterparts. Data poisoning in the continual learning setting differs from previous work from two key perspectives. First, the current research on poisoning the data streams is still under i.i.d assumptions that the model received sequential data sampled from the same distribution. Second, continual learning can be regarded as an intersection of online and offline learning. The data streams come in different tasks, and the models are trained offline on each task. These differences make poisoning attacks against continual learning an open question. Backdoor attacks against continual learning have recently been proposed to create a ``false memory'' about the targeted task~\cite{umer2020targeted, umer2022false}, called \emph{false memory formation}. The \emph{false memory formation} inserts mislabeled samples with backdoor patterns to force the neural network to create a ``false memory'' that associates the backdoor patterns with a specific task or class. Thus, samples from tasks under attack with backdoor patterns can easily fool the neural network at test time.


In this paper, we pose a more general scenario than the \emph{false memory formation} based on our previous work~\cite{huayu2022ijcnn}, that we explore the possibility of creating catastrophic forgetting intentional by data poisoning in a continual learning environment. Specifically, this work introduces poisoning attacks that significantly reduce the performance of continual learners on a target task. 
By training a continual learning model on the poisoned dataset, the model will forget the previously learned knowledge from a target task, which performance on the other tasks may seem unaffected. We first demonstrate that Label Flipping Attack \cite{Rosenfeld2020ICML} can produce significantly more damage in continual learning than offline settings. Injecting samples from the target task into upcoming tasks could produce serious catastrophic forgetting even with a small amount. Further, we propose a stealthy {\em Poisoning Attack against Continual Learning} (PACOL) derived from the label flipping attacks that do not need to change the label or inject historical data, while being difficult to detect. We evaluated the vulnerability of commonly used generative replay and regularization-based continual learning approaches using continual learning benchmarks such as Rotation-MNIST, Split-MNIST, Split-SVHN, and Split-CIFAR10. Also, we evaluate the data sanitization defense to filter out adversarial samples, and we demonstrate that the proposed attack is more difficult to detect than other attacks. Hence, we refer to PACOL's data poisoning samples as stealthy because they are challenging to identify. 

%% file: content/related_works.tex
\section{Related Work}

\subsection{Continual Learning}
We evaluate the vulnerability of continual learning in two critical and practical scenarios known as domain and task incremental learning~\cite{van2019three}. In domain incremental learning, data are sampled from distributions/domains between tasks with a fixed number of classes (i.e., the same classes are present in each task presented over time).
In contrast, task incremental learning is more challenging because the data distributions/domains and the classes differ between tasks. Thus, the data distributions change; however, the number of classes remains fixed. Further, when the tasks are learned sequentially, the labels are assigned to class IDs based on their original categories. Thus, the inference phase can predict the classes of the inputs without task IDs. 
Continual learning algorithms are generally categorized into architecture-based, replay-based, and regularization-based methods. Architecture-based methods separate the neural network into sub-networks (i.e., the sub-networks share the weights with the main network) for each task to ensure the parameters for each task have minimal effect on the other sub-networks~\cite{rusu2016progressive, aljundi2017expert}. Replay-based methods use the stored samples from previous tasks~\cite{lopez2017gradient, chaudhry2018efficient} or generative models to generate the samples of previous tasks~\cite{shin2017continual} as memories. Then, the stored or generated examples are replayed and concatenated with the current task's training to reduce the effects of catastrophic forgetting. Regularization-based approaches~\cite{kirkpatrick2017overcoming, schwarz2018progress, zenke2017continual} were proposed to address data storage and privacy problems in replay-based methods and complexity issues associated with architecture-based methods. These methods compute the importance of the learned tasks for each parameter and store the importance as a matrix. When the network is trained on new tasks, the importance matrix is used as regularization to prevent large updates to the parameter associated with old tasks~\cite{Ruvolo2013ICML}. 

Regularization-based approaches are helpful because they neither store data from previous tasks nor add more layers or nodes to the network with each incoming task. Unfortunately, their capacity to learn from challenging datasets is not ideal, as they cannot access previous or generated data. Generative replay-based methods perform better than regularization-based approaches in learning data from continuously changing distributions, while they use generative models to avoid access to historical data compared to exemplar replay-based methods. Thus, in this work, three common regularization-based algorithms, namely, Elastic Weight Consolidation (EWC)~\cite{kirkpatrick2017overcoming}, Online Elastic Weight Consolidation (online EWC)~\cite{schwarz2018progress}, and Synaptic Intelligence (SI)~\cite{zenke2017continual} are considered for simple domain incremental learning scenarios. In contrast, Deep Generative Replay (DGR)~\cite{shin2017continual} is considered for both the domain incremental learning scenarios and more complicated task incremental learning scenarios.

\subsection{Adversarial Machine Learning}
With the broad application of neural networks, the reliability and robustness issues have drawn much attention in recent years. Adversarial machine learning explores the vulnerabilities of machine learning algorithms and performs attacks to control the behavior of machine learning algorithms in two significant aspects. \emph{Evasion/exploratory} attacks~\cite{szegedy2013intriguing, goodfellow2014explaining,carlini2017towards,madry2017towards, Sadeghi2020TETCI} exploit blind spots in the models where they catastrophically misclassify the inputs with adversarial perturbations. Evasion attacks find adversarial perturbations by maximizing the loss with respect to the inputs under certain constraints. The data with even imperceptible perturbations in the direction would significantly change the feature representations in a deep network and the logit outputs. 

\emph{Poisoning/causative} attacks inject malicious data points into the training data to enforce the model converging to the wanted points of the attackers. Poisoning attacks can be broadly categorized into backdoor, targeted data poisoning, and poison availability attacks. In the backdoor attacks setting~\cite{chen2017targeted}, the adversarial samples are generated with backdoor patterns inserted into the training images to create an association between the backdoor pattern and the incorrect labels. During testing, the neural network can be fooled by inputs tagged with predefined backdoors -- or triggers -- while performing similarly to the normal model when the triggers do not activate the backdoor. Thus, backdoor attacks are regarded as intermediaries between data poisoning attacks and evasion attacks. Targeted data poisoning attacks~\cite{geiping2020witches, shafahi2018poison, zhu2019transferable} insert triggerless backdoor with only access to the training data, but cannot modify test data. Targeted data poisoning aims to cause the target samples to be misclassified to a specific class. Most works on poison availability attacks focused on reducing the accuracy in general that makes the model unusable~\cite{munoz2017towards}, and producing vanishing gradientsduring training with the modified data that make the data unexploitable~\cite{huang2021unlearnable}.

\emph{False Memory Formation} in continual learners~\cite{umer2022false, umer2020targeted} aims to control the behavior of the continual learning models by injecting backdoor attack samples. It focuses on one question whether an adversary can inject ''false memory'' on the historical task, in which the data are not shown up anymore in the current tasks. In \emph{False Memory Formation} settings, the adversary attacks the data streams by inserting the imperceptible backdoor pattern into the task's data and assigning the label of these malicious samples to incorrect classes. The continual learners being attacked are trained on several tasks with clean data and then subsequently trained on the tasks with compromised training data. The continual learners trained on the tasks with compromised training data will memorize the backdoor pattern and associate it with the incorrect label. At inference time, the adversary can fool the models into classifying the historical task's backdoor-tagged data as the desired wrong class.

%% file: content/attacks.tex
\begin{algorithm}[t]
\caption{The proposed PACOL}
\label{alg:attacks}
\textbf{Input}: model parameters: $\theta$; label flipped batches: $\tilde{\mathcal{D}}_\tau=\left \{ X_\tau,Y^{adv}_\tau \right \}$; poisoning dataset batches: $\tilde{\mathcal{D}}_{\tau+n}=\left \{ X_{\tau+n},Y_{\tau+n} \right \}$; bound of perturbation: $\epsilon$; loops: $K$; iterations: $S$; step size: $\alpha$;

\begin{algorithmic}[1]
\State \textbf{Initiate}: $s=0$; $k=0$; $\theta_0=\theta$; $X^{adv}_{\tau+n}=X_{\tau+n}$;
\While{$k \leq K$}
\State Compute label flipped gradients: 
\[
\Delta_\theta^{lf}=\nabla_\theta\mathcal{L}(f(X_\tau,\theta_k),Y_\tau^{adv})
\]
\State $s = 0$
\While{$s \leq S$}
\State Compute poisoning gradients: 
\[
\Delta_\theta^{adv}=\nabla_\theta\mathcal{L}(f(X^{adv}_{\tau+n},\theta_k),Y_{\tau+n})
\]
\State Compute cosine similarity: 
\[ 
\mathcal{H} = \text{dist}(\Delta_\theta^{lf}, \Delta_\theta^{adv}) 
\]
\State Update poisoning samples: 
\[X^{adv}_{\tau+n}=\text{clip}_{X,\epsilon}\left\{ X^{adv}_{\tau+n}+\alpha\cdot\textrm{sign}(\nabla_X\mathcal{H})\right \}
\]
\State $s=s+1$;
\EndWhile
\State Update model parameters: 
\[
\theta_{k+1} \leftarrow \text{opt-alg}_{\theta}\left \{ \mathcal{L}(f(X^{adv}_{\tau+n},\theta_{k}),Y_{\tau+n}) \right \}
\]
\State $k=k+1$
\EndWhile
\end{algorithmic}

\textbf{Output}: poisoning batches: $\tilde{\mathcal{D}}_{\tau+n}=\left \{ X^{adv}_{\tau+n},Y_{\tau+n} \right \}$;
\end{algorithm}

\section{Poisoning Attacks against Continual Learning}

This section discusses our proposed {\em Poisoning Attacks against COntinual Learning}, namely PACOL. We introduce the label-flipping attack, and the PACOL stealthy poisoning attack. We also introduce regularization-based continual learning algorithms and deep generative replay-based methods as preliminaries. The main strategy of poisoning attacks against continual learners is: Consider that a continual learner is already trained on one or several task(s), the adversary inserts a small amount (e.g., $1\%$) of malicious samples to the current task(s) (which we refer to as \emph{non-targeted task(s)}) to force the continual learner to forget the previous task(s) (which we refer to as \emph{targeted task(s)}).
In a label-flipping attack, the adversary can inject samples from the \emph{targeted task} to the \emph{non-targeted tasks} and then change the sample's label. For PACOL, the adversary has limited control over the training data that the adversary can add $\ell_\infty$-norm $\epsilon$-bounded perturbations to the data of the \emph{non-targeted tasks}. Further, the adversary performs only clean-label attacks, which means that the adversary is not permitted to change the original label of a poisoning sample.


\subsection{Regularization-based and generative replay-based continual learning}
In the continual learning setting, the model receives new pairs of training data labels $\mathcal{D}_\tau$ from a task $\tau=1,\dots,T$. The goal of the continual learner is to find the (near-)optimal parameters $\theta^*$ that minimize the empirical risk across all tasks. The objective of regularization-based continual learning approaches can be written as:
\begin{align}
    \min_\theta \underbrace{\mathcal{L}_{\theta}(\mathcal{D}_\tau)}_{\text{Risk}}+ \underbrace{\lambda\sum_i I_{\tau-1,i}(\theta_{\tau,i}-\theta_{\tau-1,i}^*)^2}_{\text{Regularization}},
\end{align}
where the regularization term penalizes changes to those parameters proportional to the importance matrix $I_{\tau-1,i}$ of the $i$th parameter calculated from the previous tasks $\mathcal{D}_{\tau-1}$. The regularization coefficient $\lambda \geq 0$ controls the \emph{stability} and
\emph{plasticity} of the model that the larger $\lambda$ results in the continual learner retaining more previous knowledge and less learning on new tasks, and vice versa. 

On the other hand, generative replay approaches, such as deep generative replay (DGR)~\cite{shin2017continual}, use generative adversarial nets (GAN)~\cite{goodfellow2014generative} trained on the tasks to generate representative pseudo-samples of the historical tasks, and use the generated samples as memories that are concatenated with the new coming tasks. At task $\tau$, DGR trains the model via the following objective:
\begin{align}
    \min_\theta \underbrace{r\cdot\mathcal{L}_{\theta_\tau}(\mathcal{D}_\tau)}_{\text{Current}}+\underbrace{(1-r)\cdot\mathcal{L}_{\theta_\tau}(\mathcal{D}_g)}_{\text{Replay}}
\end{align}
where $\mathcal{D}_g$ refers to the generated samples, and $r$ is a ratio of mixing real and generated samples, which is typically defined as $r = \frac{1}{\tau}$.

\subsection{Label-flipping attacks}
We show poisoning samples do not need to be inserted into the training data of the targeted task when the model is training on data streams. The misinformation can be injected into the models of any attacker-chosen future task. A label-flipping attack is a specific type of data poisoning attack where the adversary can change the training labels \cite{Rosenfeld2020ICML}. In offline learning settings~\cite{Xiao2012ECAI, biggio2011support}, there are two main strategies for the label-flipping attack. Adversarial label-flipping attacks aim to find the optimal label flips of the selected subset to do maximal damage to the model performance \cite{Rosenfeld2020ICML, Frederickson2018IJCNN}. Random label-flipping attacks randomly select samples from the training data at random and then flip their labels.

Label-flipping attacks can be considered the most intuitive way to erase the knowledge learned from the model. Recall that the poisoning attacks are deployed when the continual learner is first trained on a clean target dataset $\mathcal{D}_\tau$ at time $\tau$ then updated over time on several poisoned datasets $\mathcal{D}_{\tau+n}, \forall n=1,\dots, T-\tau$. Formally, we want to find the adversarial subsets denoted by $\mathcal{D}^{adv}_{\tau+n}$ through the following objective:
\begin{align}
    \max_{\mathcal{D}^{adv}_{\tau+n}} \mathcal{L}_{\theta^*}(\mathcal{D}_{\tau}), \textrm{and}\ \theta^*\in \arg\min_\theta  \sum_{n=1}^{T-\tau} \mathcal{L}_{\theta}(\mathcal{D}_{\tau+n} \cup \mathcal{D}^{adv}_{\tau+n}), 
\end{align}
where the $\mathcal{D}_{\tau}$ and $\mathcal{D}_{\tau+n}$ are not sampled from the same underlying distribution. This min-max optimization problem is difficult to solve; however, label-flipping attacks can provide an optimization-free approach to estimating a solution to this task. Consider a subset $\tilde{\mathcal{D}}_\tau$ of the \emph{targeted task} was selected, we flip the labels of the samples to the wrong categories to form the poisoning subset $\tilde{\mathcal{D}}^{adv}_\tau$. For binary classification, we flip the labels $Y_\tau\in\left \{ -1,1 \right \}$ simply by multiplying $-1$, while for multi-class classification, we flip the labels $Y_\tau\in\left \{0,\dots, n \right \}$ by the assigning the labels to another class as $Y_\tau^{adv}=(Y_\tau+z) \% (n+1)$, where $\%$ is modulo operation and $z$ is a random integer less than $n$. Thus, the label-flipping attack can be formed as:
\begin{align}
    \arg\min_\theta  \sum_{n=1}^{T-\tau} \mathcal{L}_{\theta}(\mathcal{D}_{\tau+n}) + \underbrace{\mathcal{L}_{\theta}(\tilde{\mathcal{D}}^{adv}_\tau)}_{\uparrow\mathcal{L}_{\theta}(\mathcal{D}_{\tau})},
\end{align}
It is easy to see that when the model minimizes the loss on the poisoned dataset, the error on the \emph{targeted task} increases. We adopted random label-flipping attacks here, which are already strong poisons examined by our experiments. 

\subsection{Poisoning Attack against Continual Learner}
Our goal is to derive the PACOL algorithm by starting with label-flipping attacks. Let us consider a single step in gradient descent on the dataset under label-flipping attacks without regularization and replay buffers. These assumptions allow us to simplify the problem. The model parameters after training on a task at task $\mathcal{D}_{\tau+n}$ are updated as:
\begin{align}
\label{eq:label_flip_gd}
     \theta_{\tau+1} &= \theta_{\tau}-\eta\nabla_\theta\mathcal{L}_{\theta_{\tau}}(\mathcal{D}_{\tau+1}\cup \tilde{\mathcal{D}}^{adv}_\tau) \\
     &= \theta_{\tau}-\eta\nabla_\theta\mathcal{L}(f(X_{\tau+1};\theta_{\tau}),Y_{\tau+1})-  
     \eta\underbrace{\nabla_\theta\mathcal{L}(f(X_{\tau};\theta_{\tau}),Y_{\tau}^{adv})}_{\text{Malicious}\quad \text{gradient}},
\end{align}
This expression shows that the model minimizes the loss on the current task $\mathcal{L}_{\theta}(\mathcal{D}_{\tau+n})$, and the adversarial loss $\mathcal{L}_{\theta}(\tilde{\mathcal{D}}^{adv}_\tau)$ on the target tasks. Now, we look at the PACOL setting. The cross-entropy loss is used to implement PACOL; however, our approach generalizes to other loss functions. The goal of PACOL is to find an adversarial subset $\tilde{\mathcal{D}}^{adv}_{\tau+1}$ of the current task $\mathcal{D}_{\tau+1}$ with perturbed data points $X^{adv}_{\tau+1}$ and clean labels $Y_{\tau+1}$. Then the gradient updates can be expressed as:
\begin{align}
\label{eq:adv_gd}
    \theta_{\tau+1} &= \theta_{\tau}-\eta\nabla_\theta\mathcal{L}_{\theta_{\tau}}(\mathcal{D}_{\tau+1}\cup \tilde{\mathcal{D}}^{adv}_{\tau+1}) \\
     &= \theta_{\tau}-\eta\nabla_\theta\mathcal{L}(f(X_{\tau+1};\theta_{\tau}),Y_{\tau+1})- 
     \eta\underbrace{\nabla_\theta\mathcal{L}(f(X^{adv}_{\tau+1};\theta_{\tau}),Y_{\tau+1})}_{\text{Malicious}\quad \text{gradient}},
\end{align}
where we notice that difference between Equations~\eqref{eq:label_flip_gd} and \eqref{eq:adv_gd} is the malicious gradient. Thus, the PACOL can construct the adversarial subset $\tilde{\mathcal{D}}^{adv}_{\tau+1}$ by minimizing the distance of the two malicious gradients. If the continual learning model has gradients in adversarial subset $\tilde{\mathcal{D}}^{adv}_{\tau+1}$ that are close to the label-flipped gradients subset $\tilde{\mathcal{D}}^{adv}_\tau$ then $\tilde{\mathcal{D}}^{adv}_{\tau+1}$ will have the same poisoning effect as $\tilde{\mathcal{D}}^{adv}_\tau$ for making the model forget the knowledge on the \emph{targeted task} $\mathcal{D}_\tau$. Therefore, the adversarial subset $\tilde{\mathcal{D}}^{adv}_{\tau+1}$ is found via the perturbed data points $X^{adv}_{\tau+1}$ that meet the following condition:
\begin{align}
    \nabla_\theta\mathcal{L}(f(X^{adv}_{\tau+1};\theta_{\tau}),Y_{\tau+1}) \approx \nabla_\theta\mathcal{L}(f(X_{\tau};\theta_{\tau}),Y_{\tau}^{adv}).
\end{align}
More formally, the PACOL can be defined as a simple optimization task by minimizing the distance between two gradients:
\begin{align}
    \min_{X^{adv}_{\tau+1}} \text{dist}(\nabla_\theta\mathcal{L}(f(X^{adv}_{\tau+1};\theta_{\tau}),Y_{\tau+1}), \nabla_\theta\mathcal{L}(f(X_{\tau};\theta_{\tau}),Y_{\tau}^{adv})), \nonumber
\end{align}
where $\text{dist}$ refers to the distance between two vectors. We consider two types of distance functions. The first function we consider is the $\ell_2$ distance, which is also known as Euclidean distance, is defined as:
\begin{align}
    d(p,q)=\left \| p-q \right \|^2_2, 
\end{align}
where the distance between every two distinct vectors is a positive number, while the distance from any vector to itself is zero. 
Second, we consider the negative value of cosine similarity, which is formulated as follows:
\begin{align}
    d(p,q) = -\frac{p\cdot q}{\left \| p \right \|\left \| q \right \|}=-\frac{\sum^n_{i=1}p_iq_i}{\sqrt{\sum^n_{i=1}p_i}\sqrt{\sum^n_{i=1}q_i}},
\end{align}
the cosine similarity is bounded in the interval $\left [ -1, 1 \right ]$ and is magnitude irrelevant. One advantage of the cosine similarity is that it captures the angle between the benign and adversarial perturbations, which is useful when updating the weights with the adversarial samples.

To find poisoning samples more likely to influence models during all parts of training, we propose crafting the poisoning samples in different parameter configurations along with training iterations. We describe the optimization details in Algorithm \ref{alg:attacks}. The number of loops $K$ controls how many updates the model parameters take, as we want to obtain poisoning samples that can later continually work in different training steps. The iterations $S$ contain the steps for optimizing the poisoning samples. Here we update the poisoning samples via Projected Gradient Descent (PGD)~\cite{madry2017towards}, which is restricted under the $\ell_{\infty}$-norm with bound $\epsilon$. After each PGD loop, the model parameters are updated on the poisoning samples with optimization algorithms $ \text{opt-alg}_{\theta}$. This work uses the ADAM optimizer~\cite{Kingma2015ICLR} to update the model parameters. Note that in step 7 of Algorithm \ref{alg:attacks} PACOL clips samples to remain in a feasible region (e.g., on the interval $[0,1]$ for images). 

\subsection{Adversary's knowledge and strategies}
We can define three levels of PACOL based on the knowledge about the continual learners that the adversary controls. 

\subsubsection{White-box poisoning}
We first consider a white-box attack setting, where the adversary fully knows the continual learning model. Note that a white-box setting is the worst case, since the white-box setting allows the adversary access to the data and the model parameters. The adversary has access to the continual learning model parameters $\theta_{\tau+n-1}$, the dataset for the new coming \emph{non-targeted tasks} $\mathcal{D}_{\tau+n}$, and the dataset of target \emph{targeted task} $\mathcal{D}_{\tau}$. 
There is an assumption that the adversary does not have access to the importance matrix when attacking regularization-based methods and the model when attacking generative replay-based methods. We make this assumption to keep our approach generalizable. 
Thus, the white-box attack can leverage the information from the model to craft poisoning samples. 
However, such a strategy can be challenging to implement in a real world setting because accessing the model and its parameters can be difficult. Also, white-box poisoning attacks will raise concerns about the transferability of poisoning samples, since poisoning samples may not transfer to a different model. 

\subsubsection{Gray-box and black-box poisoning}
Now we consider threat models that make weaker assumptions than white-box poisoning, which are more realistic in many settings. Beyond the limitations in the white-box attack, we assume that the adversary does not have access to the model parameters for both gray-box and black-box poisoning. Further, the adversary does not know the model architecture for black-box poisoning. Moreover, the adversary cannot gain knowledge of the exact training data of the \emph{targeted task}. However, we allow the adversary to have an auxiliary dataset $\mathcal{D}^{aux}_{\tau}$ sampled from the \emph{targeted task} for training a surrogate model $\theta^s_{\tau}$ and calculate the malicious gradients. Then, the gray-box and black-box attacks can craft poisoning samples by the surrogate model $\theta^s_{\tau}$.

%% file: content/experiments.tex
\input{content/tables/regularization}

\begin{figure}
\centering
    \subfigure[EWC]{
        \includegraphics[width=.45\textwidth]{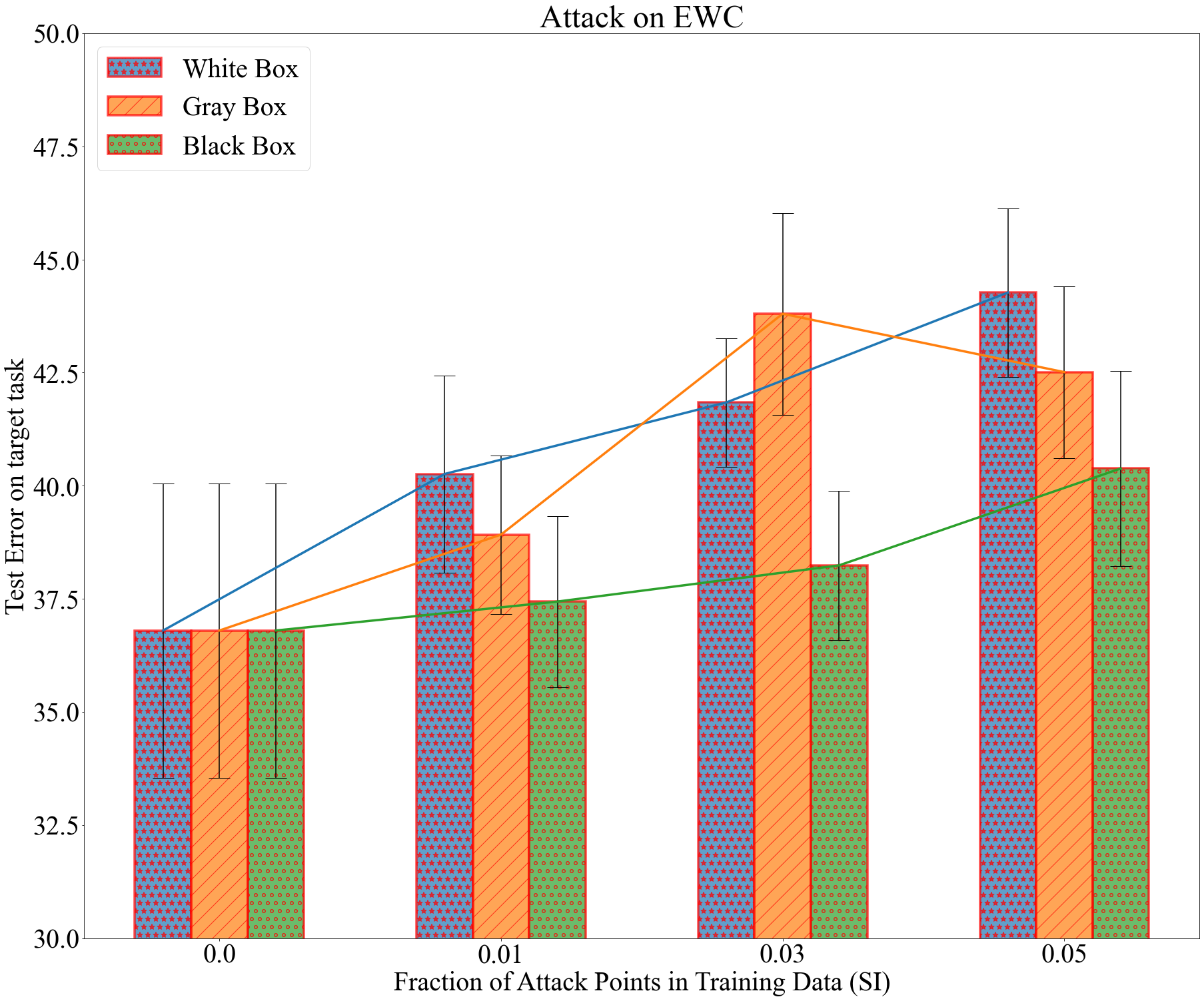}
        \label{fig:errorbars:ewc}
    }
    \subfigure[Online EWC]{
        \includegraphics[width=.45\textwidth]{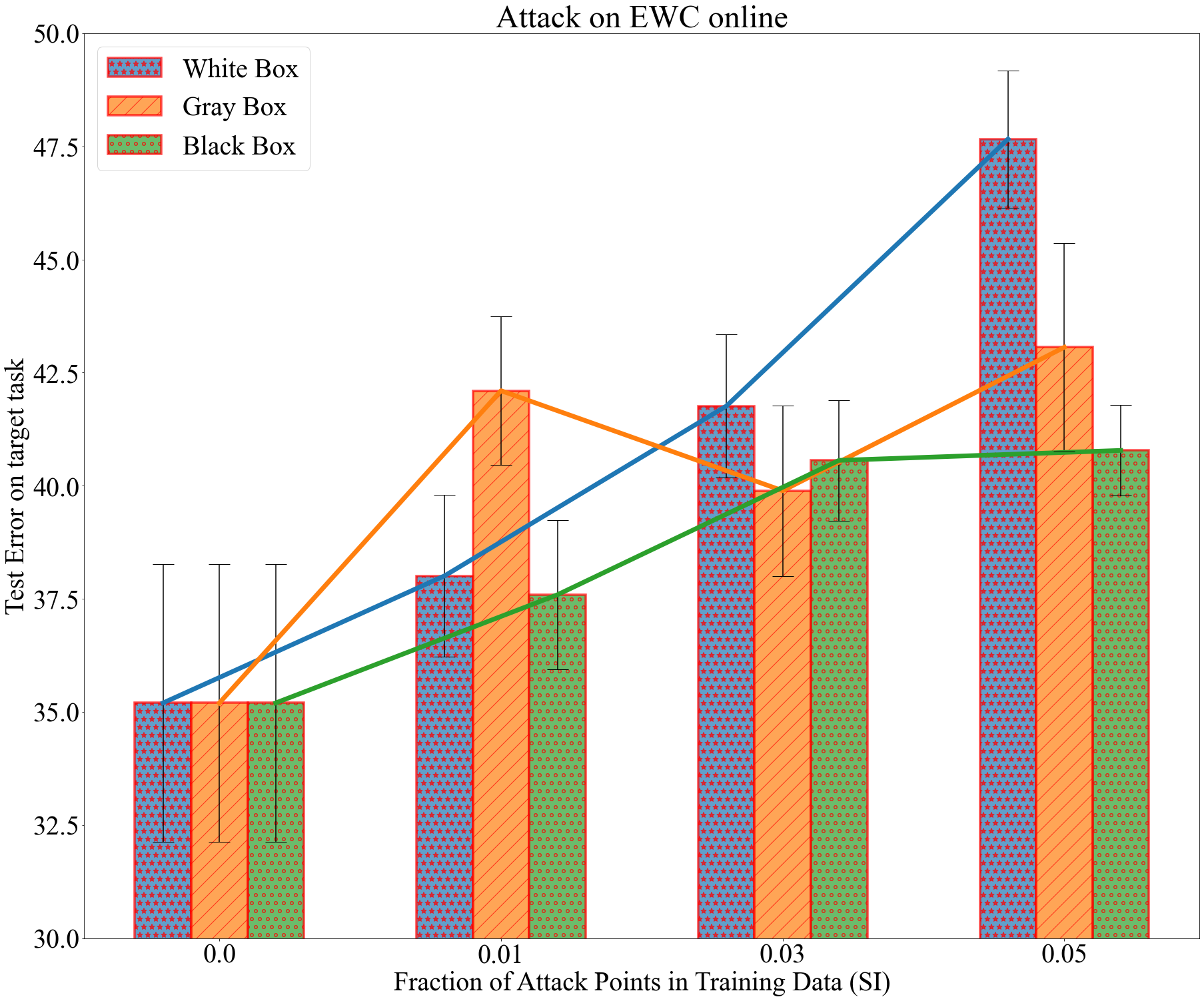}
        \label{fig:errorbars:oewc}
    } \\
    \subfigure[SI]{
        \includegraphics[width=.45\textwidth]{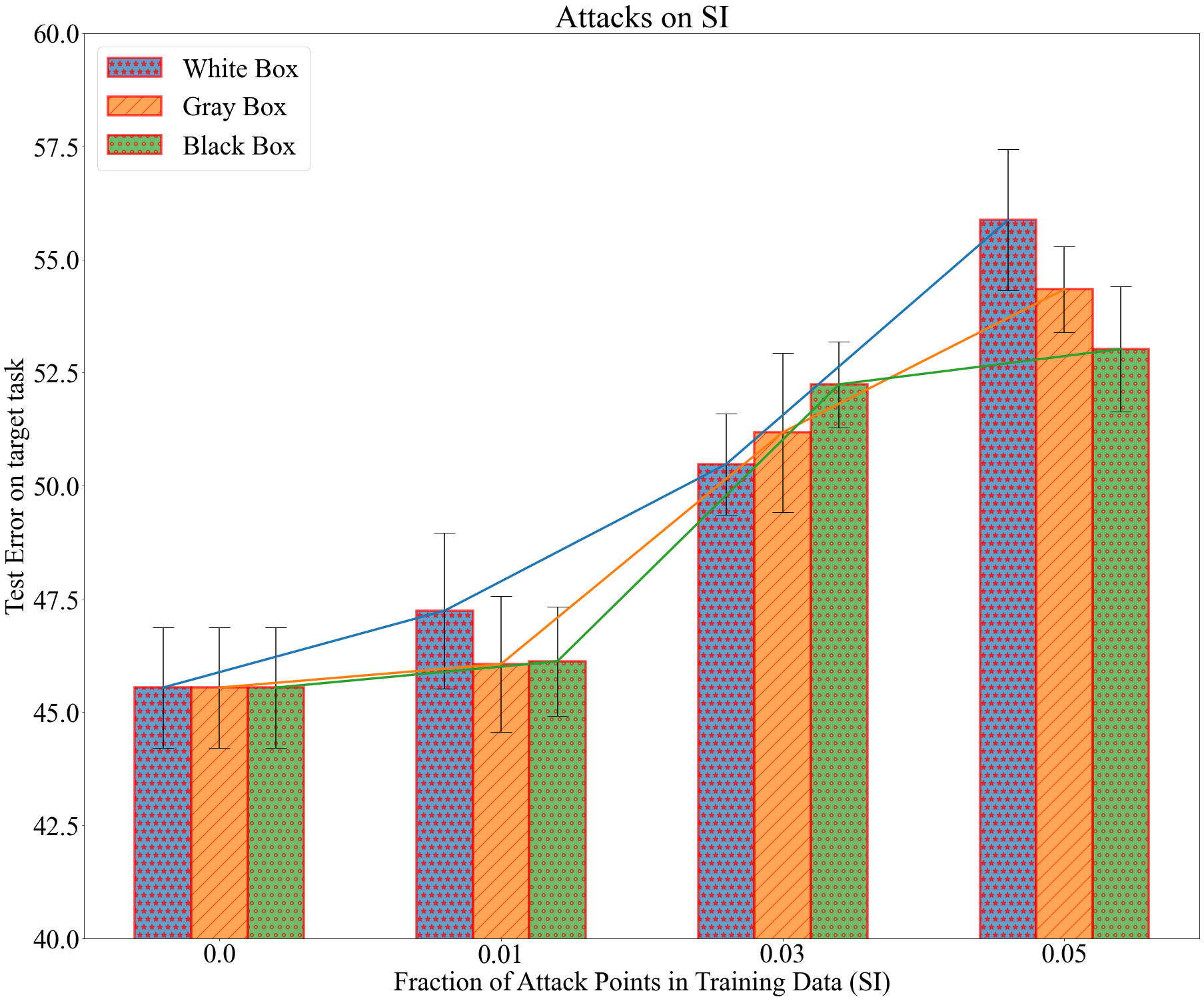}
        \label{fig:errorbars:si}
    }\bigskip
    \caption{
    Performance of EWC, Online EWC, and SI on the ROTATION MNIST dataset with different levels of poisoning attacks and different fractions of poisoned data added into the training set. The results are reported as the error, and the error bars represent a $95\%$ confidence interval. Note that the key finding here is not to compare the robustness of the three continual learning algorithms. Rather, we show that PACOL could increase the error on the \emph{targeted task} like label-flipping attacks. 
    }
    \label{fig:errorbars:reg}
\end{figure}

\section{Experiments}

\input{content/tables/dgr}

\begin{figure}
\centering
    \centering
    \subfigure[R-MNIST]{
        \includegraphics[width=.45\textwidth]{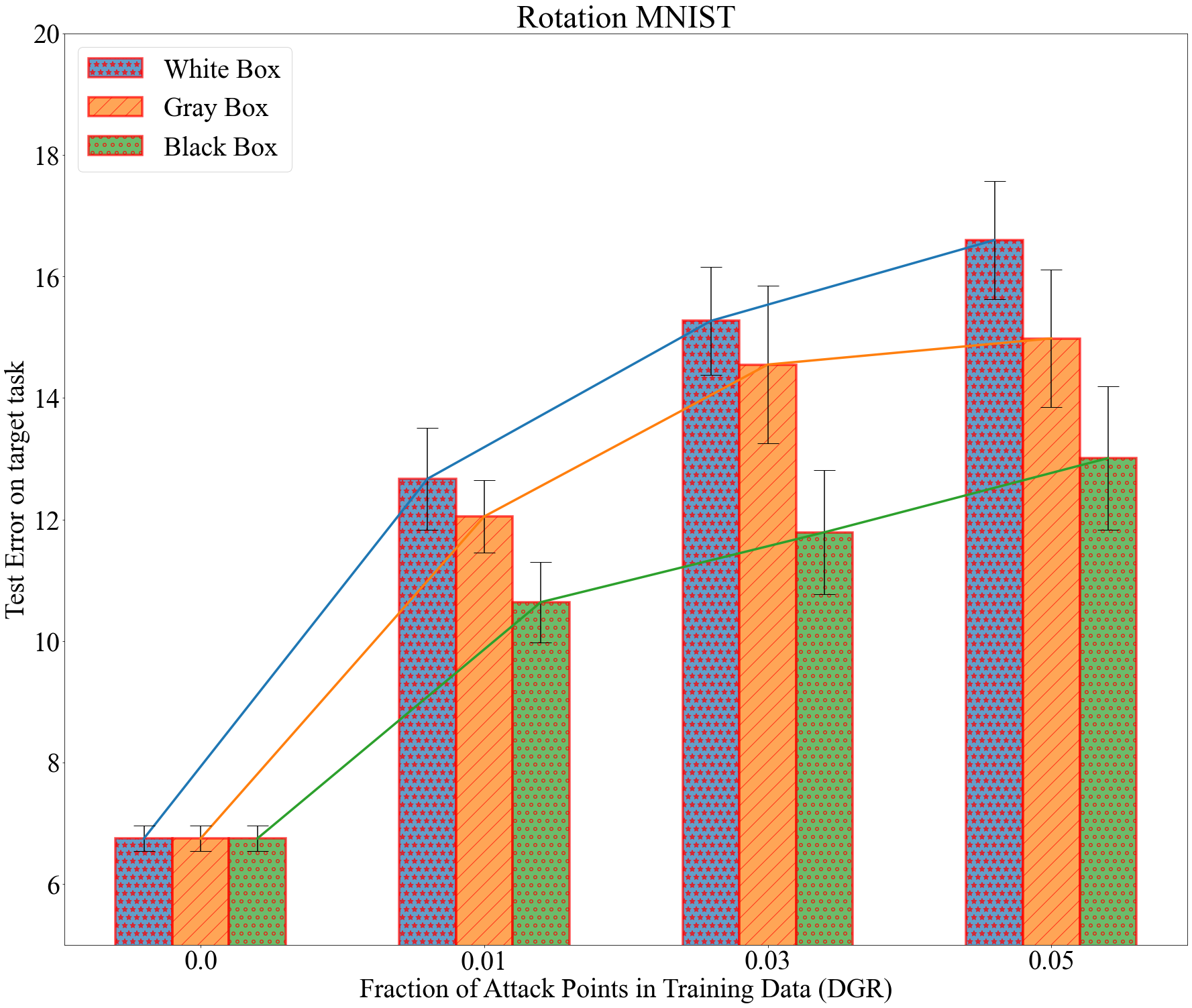}
        \label{fig:errorbars:rmnist}
    }\hfill
    \subfigure[S-MNIST]{
        \includegraphics[width=.45\textwidth]{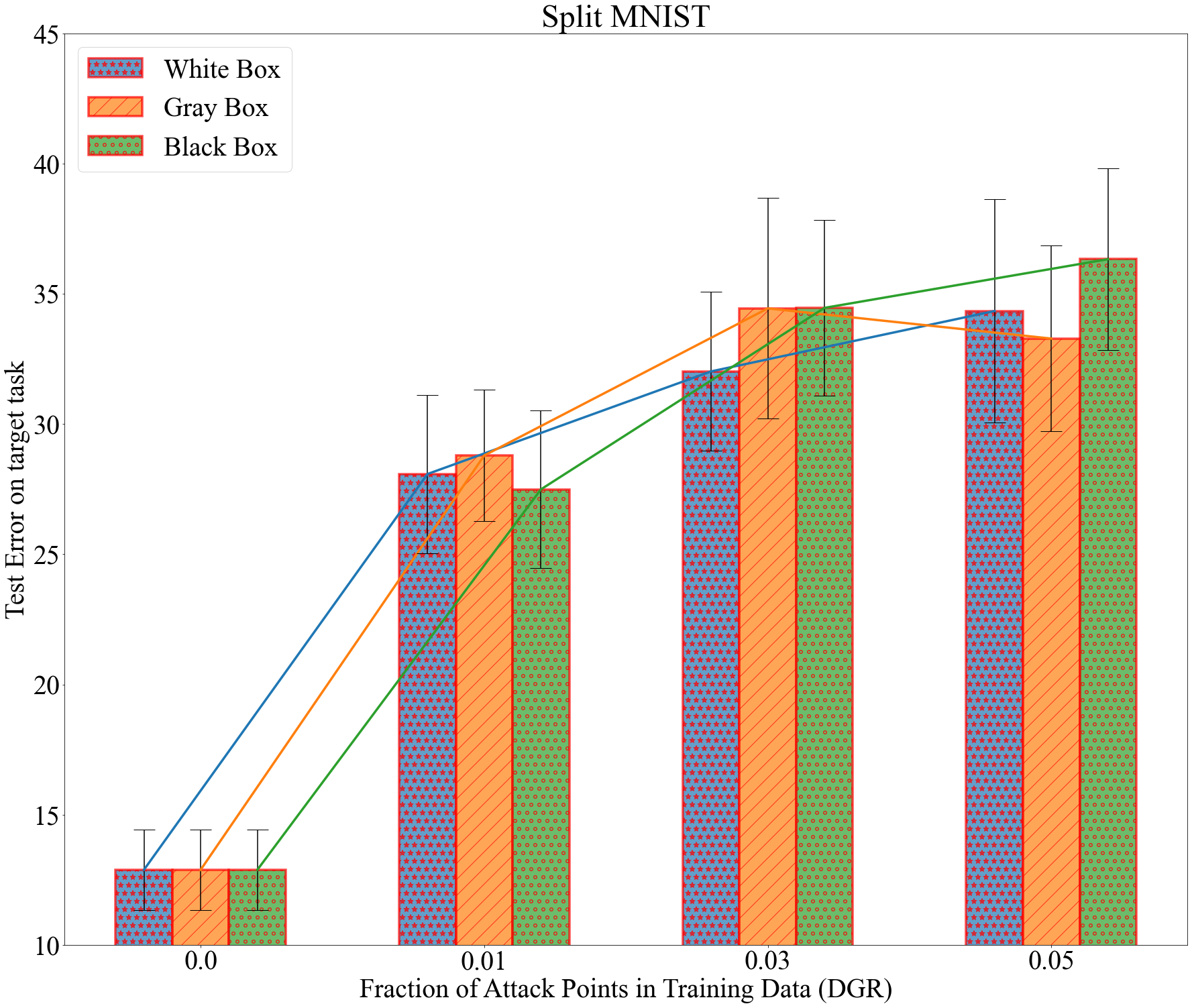}
        \label{fig:errorbars:smnist}
    }\\
    \subfigure[S-SVHN]{
        \includegraphics[width=.45\textwidth]{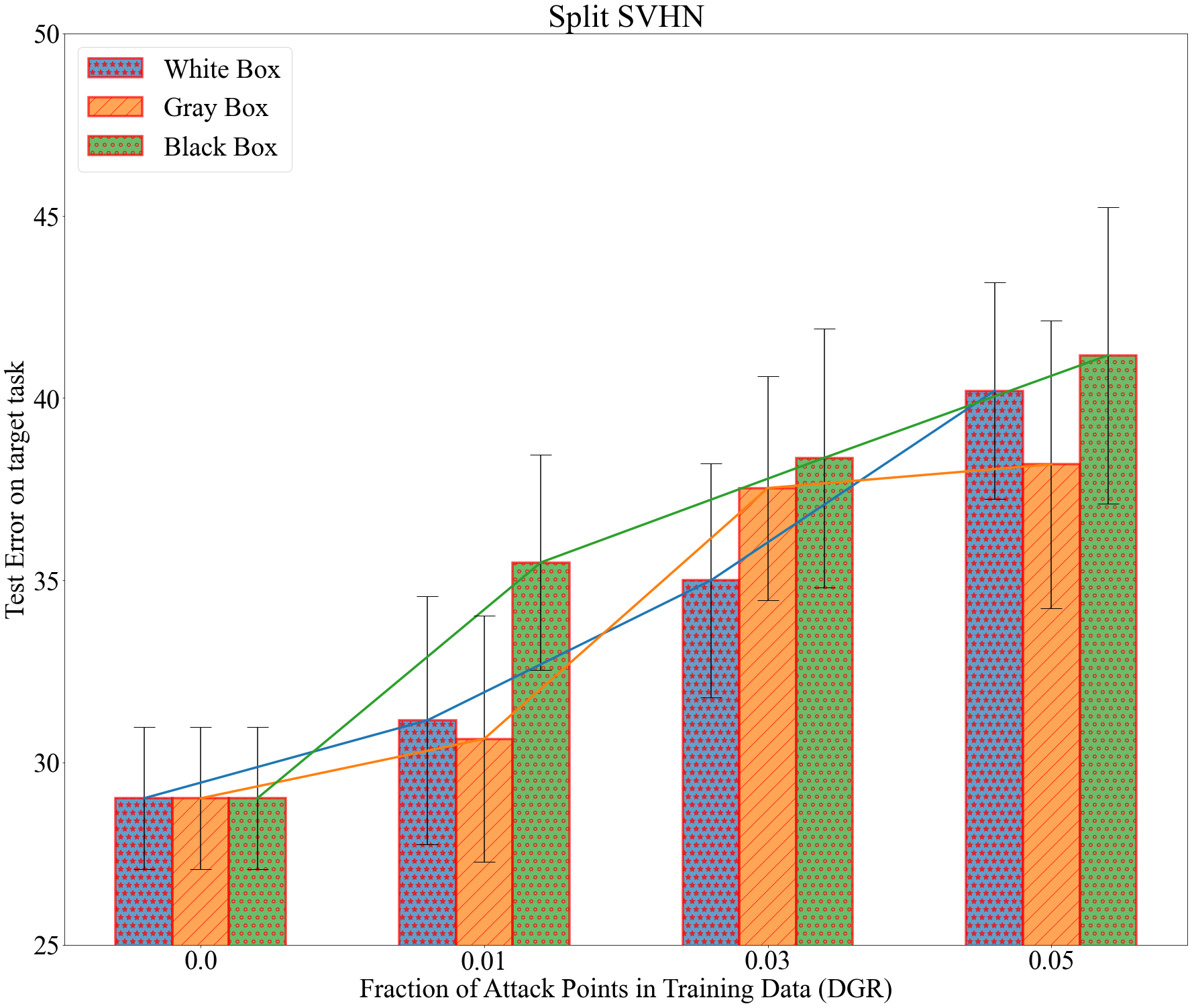}
        \label{fig:errorbars:ssvhn}
    }\hfill
    \subfigure[S-CIFAR]{
        \includegraphics[width=.45\textwidth]{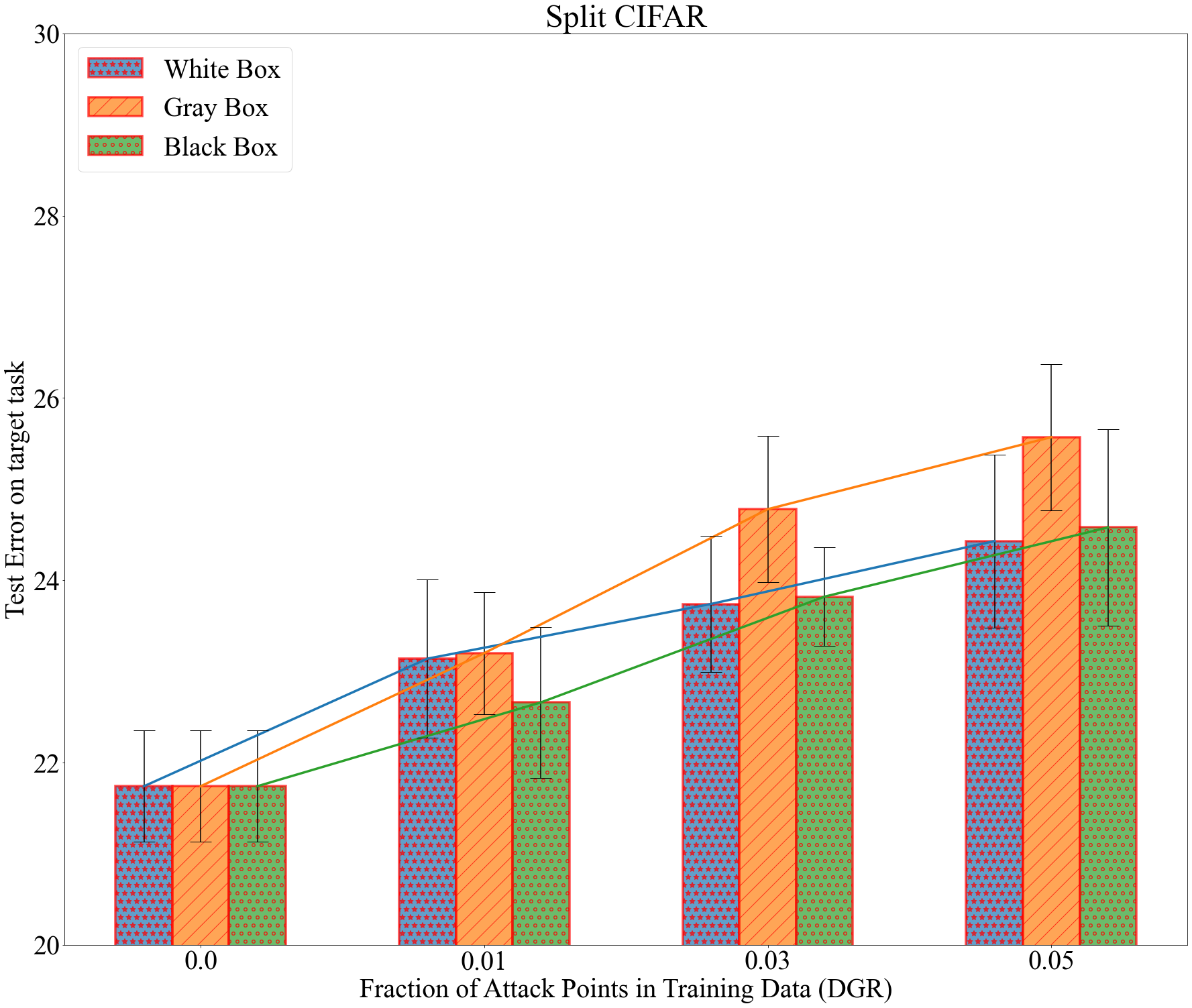}
        \label{fig:errorbars:scifar}
    }
    \caption{
    Performance of DGR across four different datasets with different levels of poisoning attacks and different fractions of poisoned data added into the training set. The results are reported as the error, and the error bars represent a $95\%$ confidence interval. We demonstrate the generality of the PACOL that it could increase the error on the \emph{targeted task} in different datasets.
    }
    \label{fig:errorbars:dgr}
\end{figure}

\input{content/tables/detection}

\begin{figure}
    \centering
    \subfigure[MNIST]{
    \includegraphics[width=.45\textwidth]{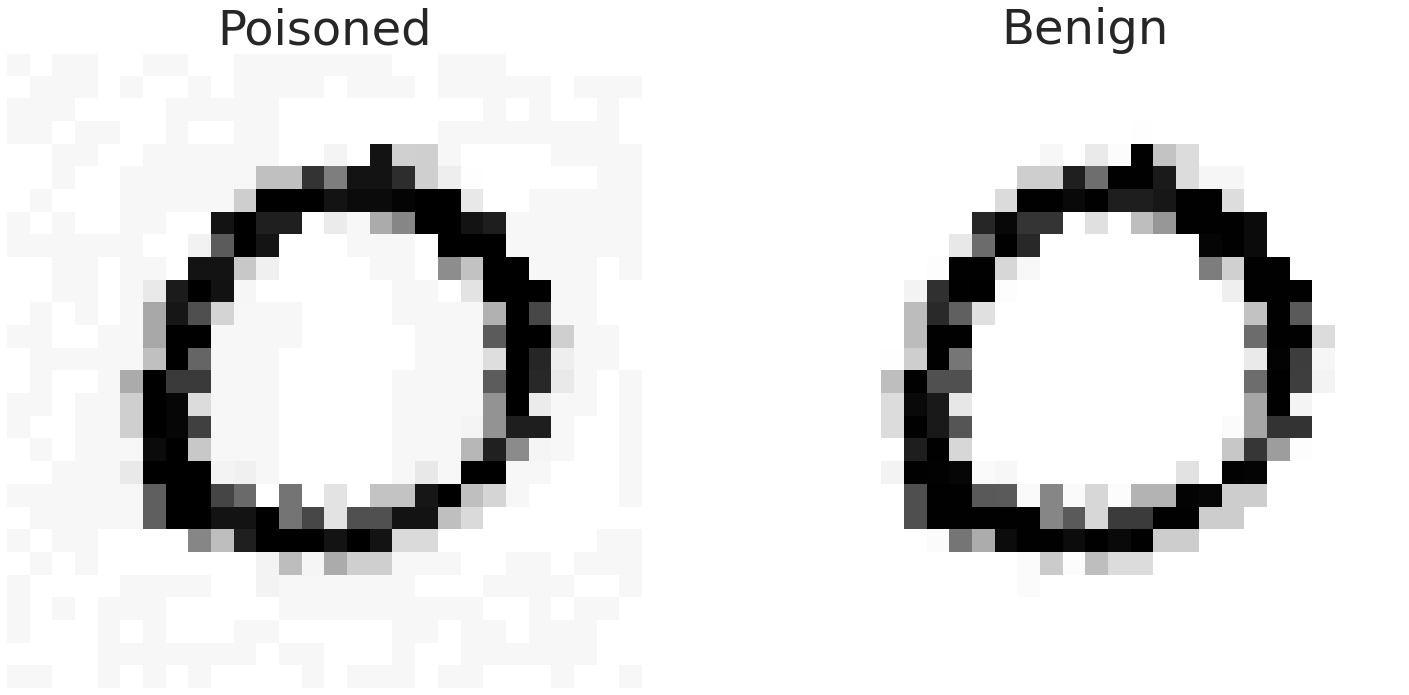}
    }
    \subfigure[CIFAR]{
    \includegraphics[width=.45\textwidth]{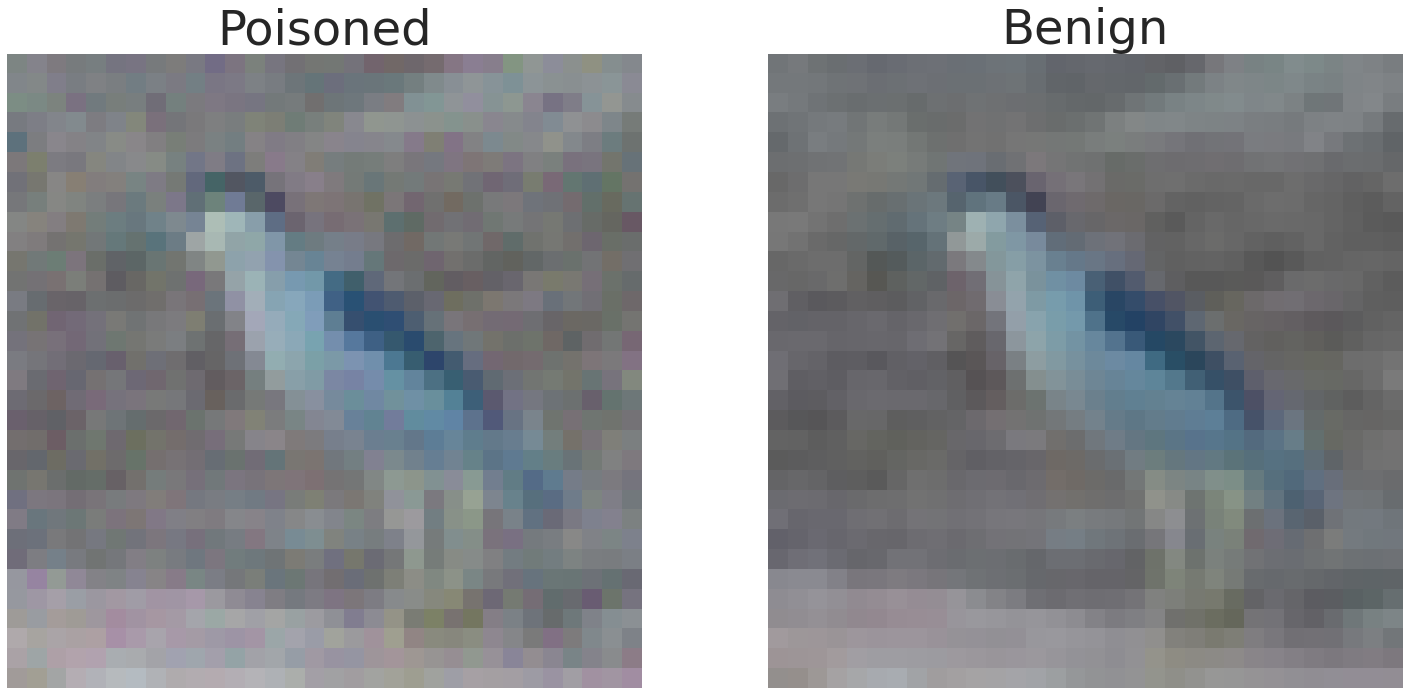}
    }
    \caption{Visual comparison between the poisoned samples and the benign samples generated by PACOL.}
    \label{fig:visualization}
\end{figure}

This section presents the experimental results for the label-flipping attack and the proposed PACOL algorithm. We first introduce our benchmarks and compare the attack methods against the three regularization-based methods, and DGR \cite{shin2017continual}. Moreover, we discuss the different defense methods for filtering out the poisoned samples generated by label-flipping attacks and the PACOL. 

\subsection{Datasets}
\label{subsec:datasets}
The continual learning algorithm's vulnerability is evaluated against label-flipping attacks and PACOL under several commonly used datasets. These datasets are variations of MNIST, SVHN, and CIFAR-10. It is worth noting that SVHN and CIFAR-10 are more challenging than MNIST. The Rotation MNIST~\cite{lopez2017gradient} (R-MNIST) dataset has a sequence of five tasks, each associated with a different rotation of the images from the original task. Each task in R-MNIST is a 10-class classification problem where the labels correspond to the digits. Thus, R-MNIST is a domain incremental learning dataset for each subsequent task that involves classification on the same ten digits. The Split MNIST~\cite{zenke2017continual} (S-MNIST) dataset also involves five tasks, each a binary classification problem. Each task is to distinguish between two digits, e.g., distinguish digits 0 and 1 in the first task, distinguish digits 2 and 3 in the second task, and so on. Split SVHN~\cite{netzer2011reading} (S-SVHN) can be regarded as the natural scene version of S-MNIST, including the house numbers in Google Street View images. Each task in S-SVHN is a binary classification problem for distinguishing digits similar to S-MNIST. Split CIFAR 10~\cite{krizhevsky2009learning} (S-CIFAR) dataset involves images from ten categories: airplanes, cars, birds, cats, deer, dogs, frogs, horses, ships, and trucks. S-CIFAR is also split into five tasks, each a binary classification problem. The adversary chooses task 1 as the \emph{targeted task} without any loss of generality. To degrade the performance of the test time on the \emph{ targeted task}, the adversary inserts malicious samples into the training data for the last two \emph{non-targeted tasks} (i.e., tasks 4 and 5). PACOL uses the $\ell_2$ distance for all MNIST variants, and negative cosine similarity as the objective for the S-SVHN and S-CIFAR datasets.

\subsection{Attacking regularization-based approaches}
\label{subsec:reg}
We first evaluate the poisoning attacks against regularization-based approaches in the R-MNIST dataset. This experiment uses a multi-layer perceptron (MLP) with 400 units and two hidden layers as the backbone network. The three continual learning algorithms trained the MLP on each task for $5,000$ iterations with batch size $128$ using the Adam optimizer~\cite{Kingma2015ICLR} with a learning rate of $0.0001$. The regularization factors are $15$ for SI, $5,000$ for EWC, and $6500$ for online EWC. 
PACOL uses Algorithm~\ref{alg:attacks} with ten loops and 40 iterations to generate poisoned samples. The adversarial perturbations are restricted under $\ell_\infty$-norm for $\epsilon=25.5/255$. The step size is $2\cdot\epsilon/T$ for attacking the three continual learning algorithms. We use the same MLP architecture with different initialization as a surrogate model for the gray-box attack and use another MLP with 100 units in the hidden layers for the black-box attack. 

We evaluate the accuracy of the clean validation sets for models trained on clean datasets and $1-5\%$ poisoned samples of tasks 4 and 5 training data as shown in Table~\ref{table:reg}. The results are averaged over ten runs, and we report the standard errors with a 95\%. 
There are several observations that we can make from this table. 
First, the white box attacks provide the largest decrease in performance compared to gray- and black-box attacks. Note that the label-flipping attack typically provides even greater performance degradation; however, these attacks are easier to detect. We discuss detection in Section \ref{subsec:defense}. 
Second, all the continual learning algorithms evaluated in this work are susceptible to PACOL attacks, demonstrating our approach's generalizability. 


We visualized the test errors on validation data of the \emph{targeted task} and the fraction of poisoned samples in the \emph{non-targeted tasks} in more straightforward error bars as shown in Figure~\ref{fig:errorbars:reg}. As the poisoning ratio increases, the error rates of the models on the \emph{targeted task} show an upward trend. 

\subsection{Attacking generative replay-based approaches}
\label{subsec:dgr}
We evaluate poisoning attacks against the DGR on the four datasets. For the MNIST variants, we use a vanilla CNN as the backbone of the continual learner and another vanilla CNN with a different structure as the surrogate model of the black-box attack. For learning on more challenging S-SVHN and S-CIFAR, we use ResNet20~\cite{he2016deep} as the backbone network and VGGnet~\cite{Simonyan2014arxiv} as the surrogate model of black-box attack. We trained a WGAN-GP~\cite{gulrajani2017improved} as the replay mechanism to mimic the past data. For each MNIST variant, we train the WGAN-GP for $8,000$ and the backbone network for $5,000$ iterations. For the natural image datasets, we train the WGAN-GP $20,000$ iterations and the backbone network as $8,000$ iterations. 

Intuitively, the replay buffers can relieve the poisoning effects that can be regarded as the \emph{correct memory}, opposite to the \emph{false memory} we inject. However, we show that the DGR is more vulnerable to data poisoning attacks than the regularization-based methods that only $1\%$ label-flipped data can significantly degrade the accuracy on \emph{targeted task}. 
Table~\ref{table:dgr} reports the results of attacks against DGR on the four datasets. The error bars are exhibited in Figure~\ref{fig:errorbars:dgr}. We generate stealthy poisoned samples with 15 loops and 40 iterations. The adversarial perturbations are restricted under $\epsilon=16/255$. We observe results similar to those we observe in Table~\ref{table:reg}. Specifically, the PACOL white box attacks provide the largest decrease in performance compared to the gray- and black box attacks. 

\vspace{-1em}

\subsection{Defense methods}
\label{subsec:defense}
Only presenting the vulnerability of the continual learning algorithms without discussing the possible solutions is incomplete. Thus, we evaluate the effectiveness of different defense methods that filter out poisoned samples. These experiments were carried out on a subset of digits 0 and 1 from the R-MNIST dataset. We use task 1 of R-MNIST as the \emph{targeted task}, and inject the poisoned samples to \emph{non-targeted task}, which is task 5 of R-MNIST. For notation, we call this new subset R-MNIST 0-1. We also evaluate the defense methods in the more complicated situation on a subset of S-CIFAR, where we use task 1 as the \emph{targeted task} and task 2 as the \emph{non-targeted task}. We poisoned 1\% of the \emph{non-targeted task} data and assigned different budgets of the defense methods to filter the poisoned samples. 

We implement the defense on different data embeddings and denote $\phi(\cdot)$ as the embedding function. We test the defense in raw data space where $\phi(x)=x$. Then, we visualize the defense using $t$-distributed stochastic neighbor embedding (t-SNE)~\cite{hinton2002stochastic}. Moreover, we use defense methods in feature space where $\phi(x)$ refers to the activations of the penultimate layer of a neural network to determine samples to filter as adversarial. In this experiment, we use the model trained on the first task as the feature extractor, which is practical in continual learning settings.  
We consider the following defense methods on the embedding space, $\phi(x)$:  
\begin{itemize}
    \item \textbf{$\ell_2$-Norm outlier defense:} The $\ell_2$-norm defense removes the fraction of points farthest in embedding space from the centroids of each class. For the data points $\left \{x_i \text{ s.t. } f(x_i)=y \right \}$ belonging to each class of label $y\in\mathcal{Y}$ with a true labeling function $f$, we compute the class centroid $c_y$ as:
        $c_y = \frac{1}{n}\sum_{x_i: f(x_i)=y} \phi(x_i)$,
    where $n$ is the number of data points in class $y$. Then we remove the data by maximizing $\left \|c_y- \phi(x_i) \right \|_2$ according to the fractions to allow the defense algorithms to remove the suspicious data points. The $\ell_2$-norm outlier defense is adapted from traditional poison defenses and is the standard in data sanitization defenses~\cite{cretu2008casting}. 
    \item \textbf{One-class SVM:} One-class SVM implements as a one-versus-all classifier, drawing a classification boundary between benign and poisoned/abnormal samples \cite{scholkopf1999support}. It maximizes the distance between the classification hyperplane and the origin in the appropriate high-dimensional feature space. Then, it takes the data points inside this hyperplane as benign and those outside of this hyperplane as outliers. 
    \item \textbf{Isolation Forest:} Isolation Forest detects anomalies using the distance between a data point and the rest of the data \cite{Liu2012ACM}. The isolation forest is an ensemble of Isolation Trees that measures the depth of the leaf node containing each data point. The outliers are categorized and isolated closer to the root node than the more clustered normal data. It exploits two distinctive properties of the outliers. First, the outliers are minority data that account for a small proportion of the entire dataset. Second, the outliers have features that are distinct from the normal samples. 
    \item \textbf{Local Outlier Factor:} Local outlier factor is build based on $k$-nearest neighbors \cite{Breunig2000ACM}. It compares the density of an instance with the density of its $k$-nearest neighbors. The local density of an outlier is expected to be much lower than the density of the $k$-nearest neighbors.
\end{itemize}

The detectability of the label-flipped samples under human censorship is higher than those with imperceptible perturbations. We demonstrate the samples with imperceptible perturbations can easily cheat human vision in Figure~\ref{fig:visualization}. We compared the defense methods in different embedding spaces by their success rate of removing the poisoned samples in Table~\ref{tab:defense_mnist} and Table~\ref{tab:defense_cifar}. When the defense methods are allowed to remove 5\% of the dataset, the $\ell_2$-Norm defense filters out nearly all label-flipped samples with raw image and deep feature embedding in the R-MNIST 0-1 subset. Further, it can remove half of the label-flipped samples with deep feature embedding in the S-CIFAR subset. However, consistent with human censorship, the $\ell_2$-Norm defense fails to detect poisoned samples with imperceptible perturbations. Surprisingly, although the One-class SVM performs worse than $\ell_2$-Norm defense in the simple case like R-MNIST 0-1 subset, it can filter out half the stealthy poisoned samples with deep feature embedding in the S-CIFAR subset. Isolation Forest defends the label-flipping attacks of the R-MNIST 0-1 subset with all embedding. Still, it cannot deal with the more complicated scenarios and stealthy poisoning attacks. Unfortunately, the Local outlier factor performs not well in any case.

%% file: content/tables/regularization.tex
\begin{table*}
\centering\footnotesize
\caption{Accuracy of the three regularization-based methods on Rotation MNIST.}
\begin{tabular}{|c|cc|c|c|c|c|c|}
\hline
\cellcolor{blue!10} Algorithm                    &  \multicolumn{2}{c|}{\cellcolor{blue!10}  Ratio (\%)}                             & \cellcolor{blue!10} Task 1         & \cellcolor{blue!10} Task 2         & \cellcolor{blue!10} Task 3         & \cellcolor{blue!10} Task 4         & \cellcolor{blue!10} Task 5         \\ \hline\hline
\multirow{13}{*}{EWC}        & \multicolumn{2}{c|}{clean}                                  & 63.2$\pm$3.25  & 75.4$\pm$3.27  & 71.06$\pm$2.69 & 70.36$\pm$1.49 & 65.85$\pm$2.49 \\ \cline{2-8} 
                             & \multicolumn{1}{c|}{\multirow{4}{*}{1\%}} & White-box      & 59.74$\pm$2.18 & 72.96$\pm$1.94 & 69.81$\pm$1.07 & 71.39$\pm$0.9  & 66.25$\pm$1.44 \\ \cline{3-8} 
                             & \multicolumn{1}{c|}{}                      & Gray-box       & 61.08$\pm$1.75 & 75.53$\pm$1.06 & 72.0$\pm$0.91  & 70.91$\pm$0.74 & 62.22$\pm$1.14 \\ \cline{3-8} 
                             & \multicolumn{1}{c|}{}                      & Black-box      & 62.56$\pm$1.89 & 75.43$\pm$1.47 & 72.97$\pm$1.27 & 70.47$\pm$1.4  & 62.37$\pm$0.99 \\ \cline{3-8} 
                             & \multicolumn{1}{c|}{}                      & Label-flipping & 60.82$\pm$1.86 & 79.38$\pm$0.64 & 74.9$\pm$1.05  & 66.17$\pm$1.54 & 60.44$\pm$1.35 \\ \cline{2-8} 
                             & \multicolumn{1}{c|}{\multirow{4}{*}{3\%}} & White-box      & 58.16$\pm$1.42 & 74.63$\pm$1.47 & 74.84$\pm$1.21 & 71.27$\pm$1.89 & 62.22$\pm$1.25 \\ \cline{3-8} 
                             & \multicolumn{1}{c|}{}                      & Gray-box       & 56.2$\pm$2.23  & 73.13$\pm$2.1  & 72.1$\pm$1.25  & 71.34$\pm$1.47 & 66.37$\pm$1.81 \\ \cline{3-8} 
                             & \multicolumn{1}{c|}{}                      & Black-box      & 61.76$\pm$1.65 & 74.25$\pm$1.91 & 72.11$\pm$1.28 & 71.89$\pm$0.67 & 62.78$\pm$0.81 \\ \cline{3-8} 
                             & \multicolumn{1}{c|}{}                      & Label-flipping & 41.07$\pm$1.59 & 73.15$\pm$2.35 & 72.57$\pm$0.99 & 73.08$\pm$1.11 & 63.24$\pm$1.48 \\ \cline{2-8} 
                             & \multicolumn{1}{c|}{\multirow{4}{*}{5\%}} & White-box      & 55.73$\pm$1.86 & 71.91$\pm$2.13 & 70.32$\pm$2.07 & 70.75$\pm$1.09 & 63.89$\pm$1.79 \\ \cline{3-8} 
                             & \multicolumn{1}{c|}{}                      & Gray-box       & 57.49$\pm$1.9  & 75.54$\pm$1.44 & 72.12$\pm$1.21 & 72.88$\pm$1.14 & 62.03$\pm$1.49 \\ \cline{3-8} 
                             & \multicolumn{1}{c|}{}                      & Black-box      & 59.62$\pm$2.15 & 74.61$\pm$2.23 & 70.18$\pm$1.99 & 71.43$\pm$1.27 & 61.16$\pm$1.9  \\ \cline{3-8} 
                             & \multicolumn{1}{c|}{}                      & Label-flipping & 33.01$\pm$0.7  & 78.9$\pm$1.38  & 73.42$\pm$1.2  & 67.83$\pm$0.64 & 60.29$\pm$0.99 \\ \hline\hline
\multirow{13}{*}{Online EWC} & \multicolumn{2}{c|}{clean}                                  & 64.8$\pm$3.07  & 72.82$\pm$2.71 & 71.13$\pm$2.27 & 73.02$\pm$1.44 & 61.86$\pm$2.74 \\ \cline{2-8} 
                             & \multicolumn{1}{c|}{\multirow{4}{*}{1\%}} & White-box      & 61.99$\pm$1.79 & 76.93$\pm$1.12 & 74.16$\pm$0.97 & 74.33$\pm$0.72 & 57.33$\pm$1.38 \\ \cline{3-8} 
                             & \multicolumn{1}{c|}{}                      & Gray-box       & 57.9$\pm$1.64  & 72.2$\pm$2.15  & 73.25$\pm$1.12 & 73.43$\pm$1.03 & 60.32$\pm$1.52 \\ \cline{3-8} 
                             & \multicolumn{1}{c|}{}                      & Black-box      & 62.41$\pm$1.65 & 75.33$\pm$1.72 & 73.14$\pm$1.21 & 71.51$\pm$0.79 & 58.28$\pm$1.54 \\ \cline{3-8} 
                             & \multicolumn{1}{c|}{}                      & Label-flipping & 54.02$\pm$1.89 & 73.74$\pm$2.52 & 73.54$\pm$1.5  & 71.95$\pm$1.3  & 61.19$\pm$2.64 \\ \cline{2-8} 
                             & \multicolumn{1}{c|}{\multirow{4}{*}{3\%}} & White-box      & 58.24$\pm$1.58 & 75.51$\pm$1.46 & 72.3$\pm$1.1   & 74.13$\pm$1.03 & 59.53$\pm$1.1  \\ \cline{3-8} 
                             & \multicolumn{1}{c|}{}                      & Gray-box       & 60.11$\pm$1.88 & 76.04$\pm$0.92 & 75.95$\pm$1.23 & 72.03$\pm$0.71 & 57.66$\pm$1.63 \\ \cline{3-8} 
                             & \multicolumn{1}{c|}{}                      & Black-box      & 59.44$\pm$1.33 & 74.89$\pm$0.91 & 73.05$\pm$1.52 & 73.64$\pm$1.05 & 57.5$\pm$1.32  \\ \cline{3-8} 
                             & \multicolumn{1}{c|}{}                      & Label-flipping & 41.91$\pm$2.16 & 73.6$\pm$2.47  & 70.98$\pm$1.75 & 72.87$\pm$0.66 & 58.03$\pm$2.7  \\ \cline{2-8} 
                             & \multicolumn{1}{c|}{\multirow{4}{*}{5\%}} & White-box      & 52.34$\pm$1.52 & 72.93$\pm$1.65 & 71.95$\pm$1.18 & 74.72$\pm$1.01 & 59.84$\pm$1.34 \\ \cline{3-8} 
                             & \multicolumn{1}{c|}{}                      & Gray-box       & 56.94$\pm$2.3  & 76.37$\pm$2.25 & 76.13$\pm$1.29 & 72.49$\pm$1.19 & 57.18$\pm$1.77 \\ \cline{3-8} 
                             & \multicolumn{1}{c|}{}                      & Black-box      & 59.22$\pm$1.0  & 75.31$\pm$1.73 & 75.08$\pm$0.88 & 74.65$\pm$0.84 & 57.23$\pm$0.89 \\ \cline{3-8} 
                             & \multicolumn{1}{c|}{}                      & Label-flipping & 32.43$\pm$0.59 & 74.72$\pm$0.72 & 74.37$\pm$0.93 & 72.92$\pm$0.6  & 55.19$\pm$1.44 \\ \hline\hline
\multirow{13}{*}{SI}         & \multicolumn{2}{c|}{clean}                                  & 54.46$\pm$1.33 & 64.27$\pm$2.36 & 68.23$\pm$2.63 & 76.57$\pm$1.42 & 75.02$\pm$1.34 \\ \cline{2-8} 
                             & \multicolumn{1}{c|}{\multirow{4}{*}{1\%}} & White-box      & 52.76$\pm$1.72 & 62.52$\pm$2.82 & 69.69$\pm$2.23 & 77.67$\pm$1.2  & 75.2$\pm$0.85  \\ \cline{3-8} 
                             & \multicolumn{1}{c|}{}                      & Gray-box       & 53.94$\pm$1.5  & 66.45$\pm$1.79 & 68.25$\pm$2.17 & 75.92$\pm$0.85 & 75.09$\pm$0.99 \\ \cline{3-8} 
                             & \multicolumn{1}{c|}{}                      & Black-box      & 53.88$\pm$1.21 & 67.26$\pm$1.82 & 68.35$\pm$1.57 & 74.95$\pm$0.66 & 73.94$\pm$0.8  \\ \cline{3-8} 
                             & \multicolumn{1}{c|}{}                      & Label-flipping & 44.67$\pm$1.81 & 63.91$\pm$2.02 & 68.93$\pm$2.18 & 78.22$\pm$0.74 & 75.38$\pm$1.57 \\ \cline{2-8} 
                             & \multicolumn{1}{c|}{\multirow{4}{*}{3\%}} & White-box      & 49.33$\pm$1.57 & 63.96$\pm$2.91 & 67.11$\pm$3.59 & 74.42$\pm$1.4  & 74.62$\pm$1.42 \\ \cline{3-8} 
                             & \multicolumn{1}{c|}{}                      & Gray-box       & 48.82$\pm$1.76 & 63.38$\pm$1.87 & 69.5$\pm$1.07  & 75.41$\pm$0.98 & 73.92$\pm$0.86 \\ \cline{3-8} 
                             & \multicolumn{1}{c|}{}                      & Black-box      & 47.76$\pm$0.95 & 59.4$\pm$2.25  & 65.5$\pm$2.18  & 77.13$\pm$0.78 & 76.18$\pm$0.85 \\ \cline{3-8} 
                             & \multicolumn{1}{c|}{}                      & Label-flipping & 29.47$\pm$0.62 & 60.83$\pm$3.58 & 65.74$\pm$3.8  & 73.94$\pm$0.83 & 76.26$\pm$1.66 \\ \cline{2-8} 
                             & \multicolumn{1}{c|}{\multirow{4}{*}{5\%}} & White-box      & 42.41$\pm$1.76 & 58.11$\pm$2.97 & 69.97$\pm$1.54 & 77.47$\pm$0.81 & 75.6$\pm$1.01  \\ \cline{3-8} 
                             & \multicolumn{1}{c|}{}                      & Gray-box       & 45.66$\pm$0.95 & 60.87$\pm$2.14 & 66.01$\pm$1.89 & 76.1$\pm$0.78  & 75.18$\pm$0.76 \\ \cline{3-8} 
                             & \multicolumn{1}{c|}{}                      & Black-box      & 46.98$\pm$1.38 & 57.24$\pm$2.45 & 65.41$\pm$2.89 & 75.09$\pm$1.35 & 76.27$\pm$1.23 \\ \cline{3-8} 
                             & \multicolumn{1}{c|}{}                      & Label-flipping & 21.4$\pm$0.27  & 62.26$\pm$2.69 & 68.1$\pm$3.77  & 72.93$\pm$2.96 & 74.52$\pm$2.03 \\ \hline
\end{tabular}
\label{table:reg}
\end{table*}

%% file: content/tables/dgr.tex
\begin{table*}
\centering\tiny
\caption{Accuracy of deep generative replay on Rotation MNIST, Split MNIST, Split SVHN, and Split CIFAR. The ratio column indicates the percentage of samples in the training datasets that are poisoned samples. }
\begin{tabular}{|c|cc|c|c|c|c|c|}
\hline
\cellcolor{blue!10} Dataset                   & \multicolumn{2}{c|}{\cellcolor{blue!10} Ratio (\%)}                          & \cellcolor{blue!10} Task 1         & \cellcolor{blue!10} Task 2         & \cellcolor{blue!10} Task 3         & \cellcolor{blue!10} Task 4         & \cellcolor{blue!10} Task 5         \\ \hline\hline
\multirow{13}{*}{R-MNIST} & \multicolumn{2}{c|}{clean}                               & 93.25$\pm$0.21 & 95.75$\pm$0.11 & 96.31$\pm$0.08 & 96.41$\pm$0.04 & 94.51$\pm$0.1  \\ \cline{2-8} 
                          & \multicolumn{1}{c|}{\multirow{4}{*}{1\%}} & White-box      & 87.33$\pm$0.84 & 94.36$\pm$0.38 & 96.1$\pm$0.42  & 97.42$\pm$0.14 & 97.03$\pm$0.07 \\ \cline{3-8} 
                          & \multicolumn{1}{c|}{}                   & Gray-box       & 87.95$\pm$0.6  & 94.5$\pm$0.41  & 96.13$\pm$0.41 & 97.48$\pm$0.11 & 97.11$\pm$0.04 \\ \cline{3-8} 
                          & \multicolumn{1}{c|}{}                   & Black-box      & 89.36$\pm$0.66 & 94.94$\pm$0.2  & 96.06$\pm$0.47 & 97.44$\pm$0.15 & 97.1$\pm$0.04  \\ \cline{3-8} 
                          & \multicolumn{1}{c|}{}                   & Label-flipping & 49.59$\pm$5.51 & 89.56$\pm$2.81 & 95.91$\pm$0.58 & 96.59$\pm$0.17 & 94.74$\pm$0.2  \\ \cline{2-8} 
                          & \multicolumn{1}{c|}{\multirow{4}{*}{3\%}} & White-box      & 84.73$\pm$0.89 & 91.84$\pm$1.16 & 94.5$\pm$0.8   & 96.98$\pm$0.19 & 97.03$\pm$0.06 \\ \cline{3-8} 
                          & \multicolumn{1}{c|}{}                   & Gray-box       & 85.45$\pm$1.3  & 92.37$\pm$1.11 & 94.8$\pm$0.9   & 97.04$\pm$0.21 & 97.13$\pm$0.08 \\ \cline{3-8} 
                          & \multicolumn{1}{c|}{}                   & Black-box      & 88.21$\pm$1.02 & 94.89$\pm$0.55 & 96.02$\pm$0.63 & 97.51$\pm$0.19 & 96.93$\pm$0.11 \\ \cline{3-8} 
                          & \multicolumn{1}{c|}{}                   & Label-flipping & 22.36$\pm$2.81 & 85.79$\pm$5.03 & 95.82$\pm$0.68 & 96.61$\pm$0.21 & 94.2$\pm$0.21  \\ \cline{2-8} 
                          & \multicolumn{1}{c|}{\multirow{4}{*}{5\%}} & White-box      & 83.4$\pm$0.97  & 94.37$\pm$0.56 & 96.25$\pm$0.53 & 97.39$\pm$0.13 & 96.78$\pm$0.11 \\ \cline{3-8} 
                          & \multicolumn{1}{c|}{}                   & Gray-box       & 85.02$\pm$1.13 & 95.23$\pm$0.38 & 96.69$\pm$0.23 & 97.55$\pm$0.04 & 96.73$\pm$0.11 \\ \cline{3-8} 
                          & \multicolumn{1}{c|}{}                   & Black-box      & 86.99$\pm$1.18 & 94.46$\pm$0.86 & 95.11$\pm$1.26 & 96.7$\pm$0.51  & 96.84$\pm$0.14 \\ \cline{3-8} 
                          & \multicolumn{1}{c|}{}                   & Label-flipping & 16.53$\pm$1.29 & 87.96$\pm$1.23 & 96.63$\pm$0.16 & 96.45$\pm$0.13 & 93.58$\pm$0.16 \\ \hline\hline
\multirow{13}{*}{S-SVHN}  & \multicolumn{2}{c|}{clean}                               & 70.98$\pm$1.95 & 68.69$\pm$4.85 & 86.77$\pm$3.26 & 97.25$\pm$0.22 & 93.53$\pm$0.27 \\ \cline{2-8} 
                          & \multicolumn{1}{c|}{\multirow{4}{*}{1\%}} & White-box      & 68.84$\pm$3.4  & 70.49$\pm$4.51 & 88.61$\pm$2.07 & 96.43$\pm$0.25 & 93.19$\pm$0.54 \\ \cline{3-8} 
                          & \multicolumn{1}{c|}{}                   & Gray-box       & 69.35$\pm$3.38 & 60.74$\pm$4.38 & 83.28$\pm$2.24 & 97.62$\pm$0.12 & 93.64$\pm$0.29 \\ \cline{3-8} 
                          & \multicolumn{1}{c|}{}                   & Black-box      & 64.51$\pm$2.95 & 67.46$\pm$3.58 & 86.93$\pm$2.62 & 96.75$\pm$0.43 & 93.75$\pm$0.23 \\ \cline{3-8} 
                          & \multicolumn{1}{c|}{}                   & Label-flipping & 35.53$\pm$4.85 & 62.4$\pm$6.33  & 84.95$\pm$4.08 & 96.37$\pm$0.38 & 93.65$\pm$0.5  \\ \cline{2-8} 
                          & \multicolumn{1}{c|}{\multirow{4}{*}{3\%}} & White-box      & 65.0$\pm$3.21  & 62.68$\pm$4.11 & 82.75$\pm$4.05 & 96.26$\pm$0.72 & 93.27$\pm$0.64 \\ \cline{3-8} 
                          & \multicolumn{1}{c|}{}                   & Gray-box       & 62.47$\pm$3.07 & 67.6$\pm$3.73  & 84.46$\pm$3.02 & 96.51$\pm$0.31 & 93.46$\pm$0.43 \\ \cline{3-8} 
                          & \multicolumn{1}{c|}{}                   & Black-box      & 61.64$\pm$3.55 & 64.63$\pm$5.52 & 90.1$\pm$1.87  & 96.96$\pm$0.34 & 93.34$\pm$0.53 \\ \cline{3-8} 
                          & \multicolumn{1}{c|}{}                   & Label-flipping & 20.07$\pm$1.98 & 62.16$\pm$5.27 & 84.67$\pm$3.26 & 96.03$\pm$0.87 & 93.82$\pm$0.48 \\ \cline{2-8} 
                          & \multicolumn{1}{c|}{\multirow{4}{*}{5\%}} & White-box      & 59.8$\pm$2.97  & 69.46$\pm$3.89 & 89.03$\pm$1.44 & 96.61$\pm$0.3  & 93.13$\pm$0.48 \\ \cline{3-8} 
                          & \multicolumn{1}{c|}{}                   & Gray-box       & 61.82$\pm$3.94 & 65.58$\pm$4.52 & 81.52$\pm$3.74 & 96.15$\pm$0.4  & 94.2$\pm$0.24  \\ \cline{3-8} 
                          & \multicolumn{1}{c|}{}                   & Black-box      & 58.83$\pm$4.06 & 67.14$\pm$3.27 & 86.76$\pm$2.62 & 96.38$\pm$0.36 & 93.82$\pm$0.36 \\ \cline{3-8} 
                          & \multicolumn{1}{c|}{}                   & Label-flipping & 14.65$\pm$1.12 & 57.77$\pm$4.49 & 84.69$\pm$1.93 & 97.02$\pm$0.35 & 93.9$\pm$0.4   \\ \hline\hline
\multirow{13}{*}{S-MNIST} & \multicolumn{2}{c|}{clean}                               & 87.12$\pm$1.54 & 95.67$\pm$0.41 & 88.06$\pm$0.77 & 99.24$\pm$0.08 & 98.05$\pm$0.09 \\ \cline{2-8} 
                          & \multicolumn{1}{c|}{\multirow{4}{*}{1\%}} & White-box      & 71.92$\pm$3.04 & 95.43$\pm$0.49 & 90.46$\pm$1.11 & 99.33$\pm$0.04 & 98.09$\pm$0.1  \\ \cline{3-8} 
                          & \multicolumn{1}{c|}{}                   & Gray-box       & 71.21$\pm$2.52 & 96.05$\pm$0.29 & 90.42$\pm$1.59 & 99.16$\pm$0.12 & 97.8$\pm$0.2   \\ \cline{3-8} 
                          & \multicolumn{1}{c|}{}                   & Black-box      & 72.51$\pm$3.02 & 96.36$\pm$0.32 & 89.4$\pm$0.89  & 99.16$\pm$0.06 & 97.96$\pm$0.1  \\ \cline{3-8} 
                          & \multicolumn{1}{c|}{}                   & Label-flipping & 22.85$\pm$1.55 & 96.83$\pm$0.37 & 92.03$\pm$1.08 & 99.02$\pm$0.09 & 97.97$\pm$0.13 \\ \cline{2-8} 
                          & \multicolumn{1}{c|}{\multirow{4}{*}{3\%}} & White-box      & 67.98$\pm$3.05 & 94.96$\pm$0.75 & 89.82$\pm$1.0  & 99.32$\pm$0.1  & 98.04$\pm$0.09 \\ \cline{3-8} 
                          & \multicolumn{1}{c|}{}                   & Gray-box       & 65.56$\pm$4.24 & 96.1$\pm$0.32  & 91.92$\pm$0.78 & 99.14$\pm$0.1  & 98.06$\pm$0.14 \\ \cline{3-8} 
                          & \multicolumn{1}{c|}{}                   & Black-box      & 65.54$\pm$3.37 & 96.22$\pm$0.49 & 91.53$\pm$0.95 & 99.24$\pm$0.08 & 98.04$\pm$0.08 \\ \cline{3-8} 
                          & \multicolumn{1}{c|}{}                   & Label-flipping & 8.9$\pm$1.02   & 94.15$\pm$0.71 & 90.09$\pm$2.07 & 99.0$\pm$0.16  & 97.92$\pm$0.06 \\ \cline{2-8} 
                          & \multicolumn{1}{c|}{\multirow{4}{*}{5\%}} & White-box      & 65.66$\pm$4.29 & 96.51$\pm$0.25 & 90.84$\pm$0.91 & 99.14$\pm$0.08 & 97.91$\pm$0.13 \\ \cline{3-8} 
                          & \multicolumn{1}{c|}{}                   & Gray-box       & 66.72$\pm$3.57 & 95.72$\pm$0.43 & 91.53$\pm$1.01 & 99.25$\pm$0.06 & 97.95$\pm$0.08 \\ \cline{3-8} 
                          & \multicolumn{1}{c|}{}                   & Black-box      & 63.67$\pm$3.5  & 96.28$\pm$0.37 & 92.42$\pm$0.92 & 99.07$\pm$0.08 & 97.99$\pm$0.11 \\ \cline{3-8} 
                          & \multicolumn{1}{c|}{}                   & Label-flipping & 4.26$\pm$0.47  & 94.41$\pm$0.8  & 91.08$\pm$2.09 & 98.58$\pm$0.19 & 97.89$\pm$0.13 \\ \hline\hline
\multirow{13}{*}{S-CIFAR} & \multicolumn{2}{c|}{clean}                               & 78.26$\pm$0.61 & 67.06$\pm$1.07 & 72.05$\pm$1.1  & 86.05$\pm$1.09 & 89.37$\pm$0.42 \\ \cline{2-8} 
                          & \multicolumn{1}{c|}{\multirow{4}{*}{1\%}} & White-box      & 76.86$\pm$0.87 & 68.44$\pm$1.15 & 73.95$\pm$1.28 & 85.51$\pm$0.46 & 88.98$\pm$0.43 \\ \cline{3-8} 
                          & \multicolumn{1}{c|}{}                   & Gray-box       & 76.8$\pm$0.67  & 65.11$\pm$1.51 & 73.69$\pm$1.69 & 87.07$\pm$0.75 & 88.58$\pm$0.16 \\ \cline{3-8} 
                          & \multicolumn{1}{c|}{}                   & Black-box      & 77.34$\pm$0.83 & 65.71$\pm$1.11 & 72.15$\pm$0.82 & 87.04$\pm$0.63 & 89.0$\pm$0.28  \\ \cline{3-8} 
                          & \multicolumn{1}{c|}{}                   & Label-flipping & 76.86$\pm$0.94 & 62.22$\pm$1.26 & 72.89$\pm$1.18 & 89.59$\pm$0.71 & 89.32$\pm$0.6  \\ \cline{2-8} 
                          & \multicolumn{1}{c|}{\multirow{4}{*}{3\%}} & White-box      & 76.26$\pm$0.75 & 63.78$\pm$0.99 & 72.51$\pm$1.45 & 87.2$\pm$0.76  & 87.94$\pm$0.44 \\ \cline{3-8} 
                          & \multicolumn{1}{c|}{}                   & Gray-box       & 75.23$\pm$0.8  & 66.97$\pm$1.02 & 73.14$\pm$0.95 & 86.4$\pm$0.59  & 89.08$\pm$0.3  \\ \cline{3-8} 
                          & \multicolumn{1}{c|}{}                   & Black-box      & 76.18$\pm$0.54 & 67.46$\pm$0.94 & 74.04$\pm$0.96 & 87.2$\pm$0.46  & 89.37$\pm$0.41 \\ \cline{3-8} 
                          & \multicolumn{1}{c|}{}                   & Label-flipping & 71.85$\pm$1.69 & 64.93$\pm$0.92 & 73.18$\pm$0.57 & 87.59$\pm$0.34 & 88.64$\pm$0.79 \\ \cline{2-8} 
                          & \multicolumn{1}{c|}{\multirow{4}{*}{5\%}} & White-box      & 75.57$\pm$0.95 & 62.65$\pm$1.6  & 70.16$\pm$1.69 & 88.07$\pm$0.61 & 87.68$\pm$0.58 \\ \cline{3-8} 
                          & \multicolumn{1}{c|}{}                   & Gray-box       & 74.44$\pm$0.8  & 66.08$\pm$1.33 & 72.8$\pm$1.07  & 87.3$\pm$0.67  & 88.27$\pm$0.26 \\ \cline{3-8} 
                          & \multicolumn{1}{c|}{}                   & Black-box      & 75.43$\pm$1.08 & 68.64$\pm$0.97 & 73.62$\pm$1.34 & 85.11$\pm$1.12 & 88.76$\pm$0.35 \\ \cline{3-8} 
                          & \multicolumn{1}{c|}{}                   & Label-flipping & 60.66$\pm$1.37 & 63.35$\pm$0.62 & 73.15$\pm$1.19 & 88.62$\pm$0.96 & 88.84$\pm$0.35 \\ \hline
\end{tabular}
\label{table:dgr}
\end{table*}


%% file: content/tables/detection.tex
\begin{sidewaystable}
\centering\footnotesize
\caption{Comparing the Success Rate (\%) of different defense methods with different data embedding on R-MNIST 0-1 subset.}
\begin{tabular}{|c|c|c|ccccc|}
\hline
\multirow{2}{*}{Methods}                 & \multirow{2}{*}{Attacks}               & \multirow{2}{*}{Embedding} & \multicolumn{5}{c|}{Buget (\%):}                                                                                                                                                      \\ \cline{4-8} 
                                         &                                        &                            & \multicolumn{1}{c|}{1}                 & \multicolumn{1}{c|}{2}                 & \multicolumn{1}{c|}{3}                 & \multicolumn{1}{c|}{4}                 & 5                 \\ \hline
\multirow{6}{*}{L2-Norm} & \multirow{3}{*}{Label-flipping} & raw data                   & \multicolumn{1}{c|}{58.57 $\pm$ 1.146} & \multicolumn{1}{c|}{77.86 $\pm$ 1.351} & \multicolumn{1}{c|}{88.81 $\pm$ 1.034} & \multicolumn{1}{c|}{95.4 $\pm$ 0.624}  & 98.49 $\pm$ 0.433 \\ \cline{3-8} 
                                         &                                        & deep feature               & \multicolumn{1}{c|}{53.89 $\pm$ 1.298} & \multicolumn{1}{c|}{73.25 $\pm$ 1.41}  & \multicolumn{1}{c|}{85.87 $\pm$ 1.164} & \multicolumn{1}{c|}{93.17 $\pm$ 0.847} & 96.43 $\pm$ 0.651 \\ \cline{3-8} 
                                         &                                        & tSNE                       & \multicolumn{1}{c|}{56.98 $\pm$ 0.876} & \multicolumn{1}{c|}{61.27 $\pm$ 1.158} & \multicolumn{1}{c|}{62.62 $\pm$ 1.298} & \multicolumn{1}{c|}{63.49 $\pm$ 1.269} & 64.37 $\pm$ 1.499 \\ \cline{2-8} 
                                         & \multirow{3}{*}{PACOL}       & raw data                   & \multicolumn{1}{c|}{1.59 $\pm$ 0.011}  & \multicolumn{1}{c|}{1.59 $\pm$ 0.021}  & \multicolumn{1}{c|}{3.97 $\pm$ 0.011}  & \multicolumn{1}{c|}{5.56 $\pm$ 0.011}  & 7.78 $\pm$ 0.106  \\ \cline{3-8} 
                                         &                                        & deep feature               & \multicolumn{1}{c|}{1.59 $\pm$ 0.011}  & \multicolumn{1}{c|}{1.59 $\pm$ 0.102}  & \multicolumn{1}{c|}{3.89 $\pm$ 0.079}  & \multicolumn{1}{c|}{5.48 $\pm$ 0.079}  & 7.86 $\pm$ 0.079  \\ \cline{3-8} 
                                         &                                        & tSNE                       & \multicolumn{1}{c|}{1.27 $\pm$ 0.103}  & \multicolumn{1}{c|}{3.89 $\pm$ 0.079}  & \multicolumn{1}{c|}{5.79 $\pm$ 0.207}  & \multicolumn{1}{c|}{7.7 $\pm$ 0.207}   & 7.94 $\pm$ 0.167  \\ \hline
\multirow{6}{*}{One-Class SVM}           & \multirow{3}{*}{Label-flipping} & raw data                   & \multicolumn{1}{c|}{49.13 $\pm$ 2.889} & \multicolumn{1}{c|}{49.13 $\pm$ 1.402} & \multicolumn{1}{c|}{49.22 $\pm$ 2.361} & \multicolumn{1}{c|}{65.63 $\pm$ 2.008} & 67.14 $\pm$ 1.136 \\ \cline{3-8} 
                                         &                                        & deep feature               & \multicolumn{1}{c|}{43.17 $\pm$ 1.791} & \multicolumn{1}{c|}{44.13 $\pm$ 2.173} & \multicolumn{1}{c|}{50.95 $\pm$ 1.569} & \multicolumn{1}{c|}{65.08 $\pm$ 1.159} & 66.03 $\pm$ 1.348 \\ \cline{3-8} 
                                         &                                        & tSNE                       & \multicolumn{1}{c|}{11.59 $\pm$ 0.582} & \multicolumn{1}{c|}{11.98 $\pm$ 0.789} & \multicolumn{1}{c|}{12.14 $\pm$ 1.236} & \multicolumn{1}{c|}{15.0 $\pm$ 0.873}  & 17.54 $\pm$ 1.402 \\ \cline{2-8} 
                                         & \multirow{3}{*}{PACOL}       & raw data                   & \multicolumn{1}{c|}{14.29 $\pm$ 0.794} & \multicolumn{1}{c|}{14.44 $\pm$ 0.746} & \multicolumn{1}{c|}{14.52 $\pm$ 0.854} & \multicolumn{1}{c|}{15.08 $\pm$ 0.909} & 17.86 $\pm$ 0.462 \\ \cline{3-8} 
                                         &                                        & deep feature               & \multicolumn{1}{c|}{13.57 $\pm$ 0.382} & \multicolumn{1}{c|}{14.21 $\pm$ 0.561} & \multicolumn{1}{c|}{15.16 $\pm$ 0.724} & \multicolumn{1}{c|}{15.56 $\pm$ 0.505} & 18.17 $\pm$ 0.522 \\ \cline{3-8} 
                                         &                                        & tSNE                       & \multicolumn{1}{c|}{6.98 $\pm$ 0.513}  & \multicolumn{1}{c|}{10.32 $\pm$ 0.659} & \multicolumn{1}{c|}{11.59 $\pm$ 0.871} & \multicolumn{1}{c|}{12.7 $\pm$ 0.939}  & 13.1 $\pm$ 0.871  \\ \hline
\multirow{6}{*}{Isolation Forest}        & \multirow{3}{*}{Label-flipping} & raw data                   & \multicolumn{1}{c|}{51.19 $\pm$ 1.253} & \multicolumn{1}{c|}{60.0 $\pm$ 1.365}  & \multicolumn{1}{c|}{66.51 $\pm$ 1.182} & \multicolumn{1}{c|}{70.87 $\pm$ 1.213} & 76.67 $\pm$ 1.281 \\ \cline{3-8} 
                                         &                                        & deep feature               & \multicolumn{1}{c|}{48.1 $\pm$ 1.046}  & \multicolumn{1}{c|}{60.56 $\pm$ 1.052} & \multicolumn{1}{c|}{70.48 $\pm$ 0.876} & \multicolumn{1}{c|}{76.75 $\pm$ 1.004} & 81.67 $\pm$ 1.271 \\ \cline{3-8} 
                                         &                                        & tSNE                       & \multicolumn{1}{c|}{91.27 $\pm$ 0.748} & \multicolumn{1}{c|}{96.67 $\pm$ 0.601} & \multicolumn{1}{c|}{97.62 $\pm$ 0.542} & \multicolumn{1}{c|}{98.02 $\pm$ 0.595} & 98.41 $\pm$ 0.516 \\ \cline{2-8} 
                                         & \multirow{3}{*}{PACOL}       & raw data                   & \multicolumn{1}{c|}{0.0 $\pm$ 0.000}   & \multicolumn{1}{c|}{2.06 $\pm$ 0.175}  & \multicolumn{1}{c|}{3.17 $\pm$ 0.118}  & \multicolumn{1}{c|}{4.76 $\pm$ 0.335}  & 6.59 $\pm$ 0.238  \\ \cline{3-8} 
                                         &                                        & deep feature               & \multicolumn{1}{c|}{0.0 $\pm$ 0.000}   & \multicolumn{1}{c|}{2.06 $\pm$ 0.175}  & \multicolumn{1}{c|}{3.49 $\pm$ 0.317}  & \multicolumn{1}{c|}{5.16 $\pm$ 0.244}  & 6.75 $\pm$ 0.319  \\ \cline{3-8} 
                                         &                                        & tSNE                       & \multicolumn{1}{c|}{1.51 $\pm$ 0.142}  & \multicolumn{1}{c|}{2.86 $\pm$ 0.175}  & \multicolumn{1}{c|}{4.44 $\pm$ 0.287}  & \multicolumn{1}{c|}{5.71 $\pm$ 0.285}  & 7.62 $\pm$ 0.359  \\ \hline
\multirow{6}{*}{Local Outlier Factor}    & \multirow{3}{*}{Label-flipping} & raw data                   & \multicolumn{1}{c|}{10.71 $\pm$ 1.092} & \multicolumn{1}{c|}{13.1 $\pm$ 1.061}  & \multicolumn{1}{c|}{15.56 $\pm$ 1.066} & \multicolumn{1}{c|}{17.86 $\pm$ 0.967} & 19.21 $\pm$ 1.023 \\ \cline{3-8} 
                                         &                                        & deep feature               & \multicolumn{1}{c|}{15.48 $\pm$ 1.324} & \multicolumn{1}{c|}{19.68 $\pm$ 1.433} & \multicolumn{1}{c|}{23.49 $\pm$ 1.557} & \multicolumn{1}{c|}{26.75 $\pm$ 1.686} & 29.84 $\pm$ 1.483 \\ \cline{3-8} 
                                         &                                        & tSNE                       & \multicolumn{1}{c|}{24.37 $\pm$ 1.435} & \multicolumn{1}{c|}{27.14 $\pm$ 1.634} & \multicolumn{1}{c|}{29.84 $\pm$ 1.548} & \multicolumn{1}{c|}{31.67 $\pm$ 1.304} & 33.1 $\pm$ 1.235  \\ \cline{2-8} 
                                         & \multirow{3}{*}{PACOL}       & raw data                   & \multicolumn{1}{c|}{0.08 $\pm$ 0.079}  & \multicolumn{1}{c|}{1.59 $\pm$ 0.058}  & \multicolumn{1}{c|}{4.44 $\pm$ 0.175}  & \multicolumn{1}{c|}{5.87 $\pm$ 0.133}  & 7.46 $\pm$ 0.131  \\ \cline{3-8} 
                                         &                                        & deep feature               & \multicolumn{1}{c|}{0.16 $\pm$ 0.106}  & \multicolumn{1}{c|}{1.59 $\pm$ 0.067}  & \multicolumn{1}{c|}{4.62 $\pm$ 0.045}  & \multicolumn{1}{c|}{5.95 $\pm$ 0.132}  & 7.56 $\pm$ 0.103  \\ \cline{3-8} 
                                         &                                        & tSNE                       & \multicolumn{1}{c|}{0.79 $\pm$ 0.089}  & \multicolumn{1}{c|}{2.62 $\pm$ 0.238}  & \multicolumn{1}{c|}{3.65 $\pm$ 0.131}  & \multicolumn{1}{c|}{5.08 $\pm$ 0.242}  & 6.43 $\pm$ 0.142  \\ \hline
\end{tabular}
\label{tab:defense_mnist}
\end{sidewaystable}

\begin{sidewaystable}
\centering\footnotesize
\caption{Comparing the Success Rate (\%) of different defense methods with different data embedding on S-CIFAR subset.}
\begin{tabular}{|c|c|c|ccccc|}
\hline
\multirow{2}{*}{Methods}              & \multirow{2}{*}{Attacks}               & \multirow{2}{*}{Embedding} & \multicolumn{5}{c|}{Buget (\%):}                                                                                                                                                                      \\ \cline{4-8} 
                                      &                                        &                            & \multicolumn{1}{c|}{1}                & \multicolumn{1}{c|}{2}                & \multicolumn{1}{c|}{3}                & \multicolumn{1}{c|}{4}                & 5                                     \\ \hline
\multirow{6}{*}{L2-Norm}              & \multirow{3}{*}{Label-flipping} & raw data                   & \multicolumn{1}{c|}{2.0 $\pm$ 0.394}  & \multicolumn{1}{c|}{4.3 $\pm$ 0.367}  & \multicolumn{1}{c|}{6.8 $\pm$ 0.389}  & \multicolumn{1}{c|}{8.9 $\pm$ 0.674}  & 10.3 $\pm$ 0.684                      \\ \cline{3-8} 
                                      &                                        & deep feature               & \multicolumn{1}{c|}{32.7 $\pm$ 0.955} & \multicolumn{1}{c|}{45.0 $\pm$ 1.398} & \multicolumn{1}{c|}{50.8 $\pm$ 1.548} & \multicolumn{1}{c|}{54.9 $\pm$ 1.362} & 57.6 $\pm$ 1.352                      \\ \cline{3-8} 
                                      &                                        & tSNE                       & \multicolumn{1}{c|}{2.0 $\pm$ 0.471}  & \multicolumn{1}{c|}{4.0 $\pm$ 0.471}  & \multicolumn{1}{c|}{6.5 $\pm$ 0.703}  & \multicolumn{1}{c|}{8.7 $\pm$ 0.448}  & 9.4 $\pm$ 0.653                       \\ \cline{2-8} 
                                      & \multirow{3}{*}{PACOL}       & raw data                   & \multicolumn{1}{c|}{0.9 $\pm$ 0.233}  & \multicolumn{1}{c|}{1.9 $\pm$ 0.348}  & \multicolumn{1}{c|}{2.6 $\pm$ 0.452}  & \multicolumn{1}{c|}{3.3 $\pm$ 0.423}  & 3.9 $\pm$ 0.348                       \\ \cline{3-8} 
                                      &                                        & deep feature               & \multicolumn{1}{c|}{0.8 $\pm$ 0.200}  & \multicolumn{1}{c|}{1.9 $\pm$ 0.100}  & \multicolumn{1}{c|}{2.5 $\pm$ 0.224}  & \multicolumn{1}{c|}{3.3 $\pm$ 0.396}  & 4.5 $\pm$ 0.543                       \\ \cline{3-8} 
                                      &                                        & tSNE                       & \multicolumn{1}{c|}{0.0 $\pm$ 0.000}  & \multicolumn{1}{c|}{0.0 $\pm$ 0.000}  & \multicolumn{1}{c|}{0.0 $\pm$ 0.000}  & \multicolumn{1}{c|}{1.9 $\pm$ 0.103}  & 4.0 $\pm$ 0.103                       \\ \hline
\multirow{6}{*}{One-Class SVM}        & \multirow{3}{*}{Label-flipping} & raw data                   & \multicolumn{1}{c|}{37.4 $\pm$ 1.771} & \multicolumn{1}{c|}{37.5 $\pm$ 1.276} & \multicolumn{1}{c|}{42.4 $\pm$ 1.035} & \multicolumn{1}{c|}{48.9 $\pm$ 1.37}  & 52.5 $\pm$ 1.869                      \\ \cline{3-8} 
                                      &                                        & deep feature               & \multicolumn{1}{c|}{24.3 $\pm$ 0.651} & \multicolumn{1}{c|}{35.7 $\pm$ 0.775} & \multicolumn{1}{c|}{43.9 $\pm$ 1.362} & \multicolumn{1}{c|}{50.4 $\pm$ 1.607} & 52.9 $\pm$ 1.709                      \\ \cline{3-8} 
                                      &                                        & tSNE                       & \multicolumn{1}{c|}{10.9 $\pm$ 0.96}  & \multicolumn{1}{c|}{14.7 $\pm$ 0.932} & \multicolumn{1}{c|}{16.4 $\pm$ 1.046} & \multicolumn{1}{c|}{16.6 $\pm$ 1.543} & 17.9 $\pm$ 1.09                       \\ \cline{2-8} 
                                      & \multirow{3}{*}{PACOL}       & raw data                   & \multicolumn{1}{l|}{9.0 $\pm$ 1.043}  & \multicolumn{1}{l|}{13.7 $\pm$ 0.895} & \multicolumn{1}{l|}{15.5 $\pm$ 1.593} & \multicolumn{1}{l|}{16.7 $\pm$ 1.023} & \multicolumn{1}{l|}{16.9 $\pm$ 0.809} \\ \cline{3-8} 
                                      &                                        & deep feature               & \multicolumn{1}{c|}{33.7 $\pm$ 0.367} & \multicolumn{1}{c|}{38.9 $\pm$ 0.547} & \multicolumn{1}{c|}{47.2 $\pm$ 0.757} & \multicolumn{1}{c|}{50.6 $\pm$ 0.601} & 57.7 $\pm$ 0.597                      \\ \cline{3-8} 
                                      &                                        & tSNE                       & \multicolumn{1}{c|}{0.0 $\pm$ 0.000}  & \multicolumn{1}{c|}{0.0 $\pm$ 0.000}  & \multicolumn{1}{c|}{0.8 $\pm$ 0.133}  & \multicolumn{1}{c|}{1.0 $\pm$ 0.000}  & 2.3 $\pm$ 0.153                       \\ \hline
\multirow{6}{*}{Isolation Forest}     & \multirow{3}{*}{Label-flipping} & raw data                   & \multicolumn{1}{c|}{2.6 $\pm$ 0.306}  & \multicolumn{1}{c|}{5.4 $\pm$ 0.653}  & \multicolumn{1}{c|}{7.8 $\pm$ 0.663}  & \multicolumn{1}{c|}{9.5 $\pm$ 0.847}  & 11.6 $\pm$ 0.833                      \\ \cline{3-8} 
                                      &                                        & deep feature               & \multicolumn{1}{c|}{31.3 $\pm$ 1.136} & \multicolumn{1}{c|}{42.7 $\pm$ 1.184} & \multicolumn{1}{c|}{50.4 $\pm$ 1.462} & \multicolumn{1}{c|}{55.5 $\pm$ 1.544} & 59.4 $\pm$ 1.401                      \\ \cline{3-8} 
                                      &                                        & tSNE                       & \multicolumn{1}{c|}{1.5 $\pm$ 0.543}  & \multicolumn{1}{c|}{3.7 $\pm$ 0.803}  & \multicolumn{1}{c|}{5.4 $\pm$ 0.957}  & \multicolumn{1}{c|}{8.4 $\pm$ 1.176}  & 10.3 $\pm$ 1.136                      \\ \cline{2-8} 
                                      & \multirow{3}{*}{PACOL}       & raw data                   & \multicolumn{1}{c|}{0.0 $\pm$ 0.000}  & \multicolumn{1}{c|}{0.2 $\pm$ 0.133}  & \multicolumn{1}{c|}{1.7 $\pm$ 0.213}  & \multicolumn{1}{c|}{2.8 $\pm$ 0.200}  & 4.0 $\pm$ 0.211                       \\ \cline{3-8} 
                                      &                                        & deep feature               & \multicolumn{1}{c|}{0.5 $\pm$ 0.224}  & \multicolumn{1}{c|}{0.6 $\pm$ 0.221}  & \multicolumn{1}{c|}{1.1 $\pm$ 0.233}  & \multicolumn{1}{c|}{1.7 $\pm$ 0.26}   & 2.6 $\pm$ 0.267                       \\ \cline{3-8} 
                                      &                                        & tSNE                       & \multicolumn{1}{c|}{0.0 $\pm$ 0.000}  & \multicolumn{1}{c|}{0.6 $\pm$ 0.221}  & \multicolumn{1}{c|}{1.3 $\pm$ 0.396}  & \multicolumn{1}{c|}{2.8 $\pm$ 0.327}  & 3.7 $\pm$ 0.448                       \\ \hline
\multirow{6}{*}{Local Outlier Factor} & \multirow{3}{*}{Label-flipping} & raw data                   & \multicolumn{1}{c|}{3.8 $\pm$ 0.442}  & \multicolumn{1}{c|}{6.3 $\pm$ 0.684}  & \multicolumn{1}{c|}{10.2 $\pm$ 0.68}  & \multicolumn{1}{c|}{13.7 $\pm$ 0.761} & 16.3 $\pm$ 0.844                      \\ \cline{3-8} 
                                      &                                        & deep feature               & \multicolumn{1}{c|}{5.3 $\pm$ 0.831}  & \multicolumn{1}{c|}{8.9 $\pm$ 1.178}  & \multicolumn{1}{c|}{12.7 $\pm$ 1.476} & \multicolumn{1}{c|}{14.3 $\pm$ 1.469} & 16.5 $\pm$ 1.258                      \\ \cline{3-8} 
                                      &                                        & tSNE                       & \multicolumn{1}{c|}{2.5 $\pm$ 0.563}  & \multicolumn{1}{c|}{4.2 $\pm$ 0.786}  & \multicolumn{1}{c|}{6.2 $\pm$ 0.952}  & \multicolumn{1}{c|}{8.6 $\pm$ 1.077}  & 10.1 $\pm$ 1.215                      \\ \cline{2-8} 
                                      & \multirow{3}{*}{PACOL}       & raw data                   & \multicolumn{1}{c|}{0.0 $\pm$ 0.000}  & \multicolumn{1}{c|}{1.0 $\pm$ 0.001}  & \multicolumn{1}{c|}{2.0 $\pm$ 0.003}  & \multicolumn{1}{c|}{2.0 $\pm$ 0.003}  & 2.0 $\pm$ 0.003                       \\ \cline{3-8} 
                                      &                                        & deep feature               & \multicolumn{1}{c|}{0.2 $\pm$ 0.133}  & \multicolumn{1}{c|}{1.0 $\pm$ 0.258}  & \multicolumn{1}{c|}{1.5 $\pm$ 0.269}  & \multicolumn{1}{c|}{2.0 $\pm$ 0.211}  & 2.9 $\pm$ 0.433                       \\ \cline{3-8} 
                                      &                                        & tSNE                       & \multicolumn{1}{c|}{0.0 $\pm$ 0.000}  & \multicolumn{1}{c|}{2.6 $\pm$ 0.163}  & \multicolumn{1}{c|}{4.0 $\pm$ 0.001}  & \multicolumn{1}{c|}{4.0 $\pm$ 0.001}  & 4.0 $\pm$ 0.001                       \\ \hline
\end{tabular}
\label{tab:defense_cifar}
\end{sidewaystable}

%% file: content/Conclusion.tex
\section{Conclusion}

Exploring continual machine learning with adversaries is more vital than ever. The continual evolution of threats and the rapid pace of change necessitate that machine learning systems not only learn from past experiences, but also adapt in real-time to new challenges. Continual learning is the pathway to this adaptive, robust intelligence. But to truly harness its potential, we must factor in adversaries -- those entities that seek to compromise our systems and misuse our technology. Adversarial tactics are always evolving, exploiting vulnerabilities in ways that we can hardly predict. Therefore, our machine learning systems must be robust, flexible, and continually learning not just from benign data, but also from these adversarial interactions. By exploring and understanding adversarial impacts in continual learning, we empower our systems to adapt, withstand, and predict potential threats, paving the way for a safer, more secure future.
This work presented a new type of data poisoning attack that exposes the vulnerability of the current adversary-agnostic continual learning algorithms to data poisoning attacks. We show that an adversary can force continual learners to forget the knowledge on a specific task with malicious samples. We first show that the label-flipping attacks can significantly reduce the model accuracy on the previously learned \emph{targeted tasks}, and then we derive the PACOL with less detectability. The poison samples erase the memory of the continual learners, making the continual learners forget the knowledge of the \emph{targeted tasks}. In addition to the presentation of data poisoning attack methods, the primary purpose of this work is to raise the community's awareness to focus on robust learning algorithms against adversarial threats in continual learning settings. In particular, the proposed approach has ethical implications because adversarial data can be used to target forgetting in a way that violates fairness in machine learning \cite{Verma2018FAT}. Therefore, we urge that care be taken when these models and techniques are used in practice, and the experiments in this work are limited to specific types of concept forgetting.  

Our future work includes developing data-, algorithmic- and architecture-level strategies to ensure adversarial robustness for continual machine learning models. Another area of critical investigation is a formal benchmark or adversaries in continual and lifelong learning environments. 